\documentclass[runningheads]{llncs}
\usepackage{graphicx}
\usepackage{tikz}
\usepackage{comment}
\usepackage{amsmath,amssymb} % define this before the line numbering.
\usepackage{color}

\usepackage{amsmath}
\usepackage{amssymb}
\usepackage{booktabs}
\usepackage{makecell}
\usepackage{multicol,multirow}
\usepackage{microtype}
\usepackage{enumitem}
\usepackage{wrapfig}
\usepackage[misc]{ifsym}

\usepackage{enumitem}
\usepackage{algorithm}
\usepackage{algorithmic}
\usepackage{alltt}
\usepackage[accsupp]{axessibility}  % Improves PDF readability for those with disabilities.

\def\0{{\bf 0}}
\def\1{{\bf 1}}

\def\ie{\text{i.e.}}
\def\eg{\text{e.g.}}

\def\eg{\emph{e.g.}} 

\def\ie{\emph{i.e.}}

\definecolor{lineblue}{RGB}{91, 155, 213}
\definecolor{linered}{RGB}{192, 0, 0}
\definecolor{OliveGreen}{RGB}{34, 139, 34}
\definecolor{Black}{RGB}{0, 0, 0}

\begin{document}

\pagestyle{headings}
\mainmatter
\def\ECCVSubNumber{65}  % Insert your submission number here

\title{Discriminability-Transferability Trade-Off:\\ An Information-Theoretic Perspective} % Replace with your title

% \begin{comment}
\titlerunning{Discriminability-Transferability Trade-Off}
\author{Quan Cui\inst{1,2}$^{\dagger}$ \and
Bingchen Zhao\inst{1,3}$^{\dagger}$ \and
Zhao-Min Chen\inst{1,4} \and
Borui Zhao\inst{1} \\
Renjie Song\inst{1}\Letter \and
Boyan Zhou\inst{5} \and
Jiajun Liang\inst{1} \and
Osamu Yoshie\inst{2}
}
\authorrunning{Q. Cui and B. Zhao et al.}
\institute{$^{\text{1}}$~MEGVII Technology~~$^{\text{2}}$~Waseda University~~$^{\text{3}}$~University of Edinburgh\\ $^{\text{4}}$~Wenzhou University~~$^{\text{5}}$~ByteDance\\
\email{cui-quan@toki.waseda.jp, zhaobc.gm@gmail.com, songrenjie@megvii.com}}
% \end{comment}
%******************
\maketitle
{\let\thefootnote\relax\footnote{{\text{$^{\dagger}$} denotes equal contributions, and \text{\Letter} denotes the corresponding author.}}}

\begin{abstract}
This work simultaneously considers the discriminability and transferability properties of deep representations in the typical supervised learning task, \ie, image classification. By a comprehensive temporal analysis, we observe a trade-off between these two properties. The discriminability keeps increasing with the training progressing while the transferability intensely diminishes in the later training period. 
From the perspective of information-bottleneck theory, we reveal that the incompatibility between discriminability and transferability is attributed to the over-compression of input information. More importantly, we investigate why and how the InfoNCE loss can alleviate the over-compression, and further present a learning framework, named contrastive temporal coding~(CTC), to counteract the over-compression and alleviate the incompatibility.
Extensive experiments validate that CTC successfully mitigates the incompatibility, yielding discriminative and transferable representations. Noticeable improvements are achieved on the image classification task and challenging transfer learning tasks. We hope that this work will raise the significance of the transferability property in the conventional supervised learning setting.

\keywords{Information-Bottleneck Theory, Representation Learning, Discriminability, Transferability, Contrastive Learning}
\end{abstract}

\section{Introduction}

% --------- what we want to do: analysis the learning process and the three properties of representations
In recent decades, great progress has been achieved on learning discriminative representations with deep neural networks in a supervised learning manner. Advanced by such powerful deep representations, performances of many real-world computer vision applications are remarkably improved, \eg, visual categorization\cite{resnet,senet,triplet}, object detection~\cite{mask,faster,densecon}, and semantic segmentation~\cite{fcn,pspnet}. However, the mainstream supervised learning works concentrate on pursuing more discriminative representations, which deserves the most attention for direct effects on model performances. Except for the discriminability, transferability should also be taken into considerations in the conventional supervised learning classification task, which is preferred for many downstream tasks~\cite{simclr,moco,instancedis,rethinkingpretrain2,improving} yet neglected in the conventional supervised learning setting.

% --------- what properties are discussed.
%Properties considered in this work are detailed as preliminary. 
%(1)~The \textit{discriminability} denotes whether representations are distinguishable on validation/test sets. 
%%(2)~The \textit{robustness} represents the ability of representations to generalize on corrupted or perturbed inputs. 
%(2)~The \textit{transferability} measures performances on transferring representations to unseen domains or other tasks. 
%This work concentrates on learning representations that can simultaneously be discriminative and transferable in the supervised learning manner. 

\begin{figure}[t]
%%\vspace{-15pt}
  \begin{minipage}[c]{0.4\textwidth}
    \includegraphics[width=1.0\linewidth]{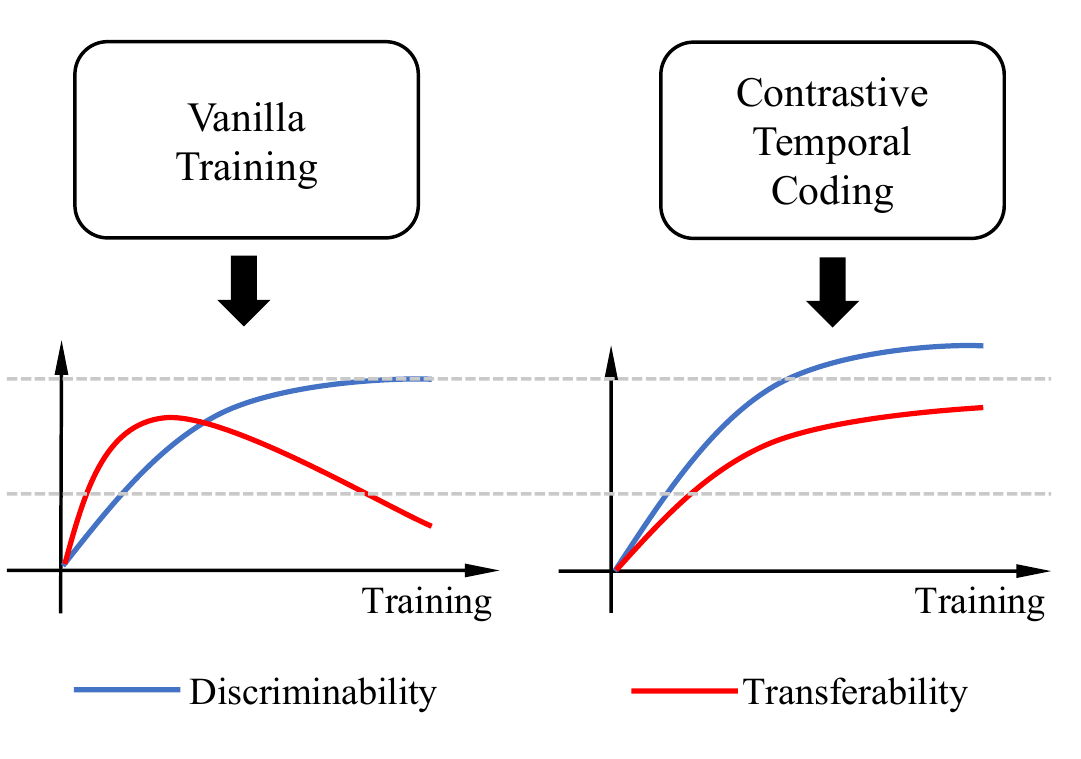}
  \end{minipage} \quad
  \begin{minipage}[c]{0.55\textwidth}
    \caption{
       Considered properties of deep representations. In a vanilla training, we reveal that the discriminability keeps increasing while the transferability first climbs to a peak and then gradually decreases. To address this issue, we propose the contrastive temporal coding, which successfully alleviates the incompatibility between transferability and discriminability.
    } \label{fig:introduction}
  \end{minipage}
% %\vspace{-15pt}
\end{figure}

% --------- how to analysis these properties, and the result of the analysis.
We start this work by investigating the correlation between discriminability and transferability properties of representations in the entire supervised training process. Concretely, in the training process of a deep classification model, we extract representations from each training epoch and respectively assess their discriminability and transferability. Conclusions are generally illustrated in the left part of Figure~\ref{fig:introduction}. We observe that the discriminability keeps getting better, while the transferability intensely diminishes in the later training. It reveals that representations can hardly be discriminative and transferable at the same time in the conventional supervised learning setting, \ie, these two properties could be incompatible. Nevertheless, high-quality representations are expected to possess both properties, and we suppose that the learning mechanism underlying the conventional supervised learning leads to the trade-off. 

%\begin{wrapfigure}{r}{0.5\linewidth}
%%\vspace{-20pt}
%\begin{center}
%\includegraphics[width=1.0\linewidth]{figure/introduction}
%\end{center}
%%%\vspace{-10pt}
%\caption{Considered properties of deep representations. In a vanilla training, we reveal that the discriminability keeps increasing while the transferability first climbs to a peak and then gradually decreases. To address this issue, we propose the contrastive temporal coding, which successfully alleviates the incompatibility between transferability and discriminability.}
%%\vspace{-20pt}
%\label{fig:introduction}
%\end{wrapfigure}

%\begin{figure}[t]
%%%\vspace{-10pt}
%\floatbox[{\capbeside\thisfloatsetup{capbesideposition={right,top},capbesidewidth=5cm}}]{figure}[\FBwidth]
%{\caption{Considered properties of deep representations. In a vanilla training, we reveal that the discriminability keeps increasing while the transferability first climbs to a peak and then gradually decreases. To address this issue, we propose the contrastive temporal coding, which successfully alleviates the incompatibility between transferability and discriminability.}\label{fig:introduction}}
%{\includegraphics[width=0.5\textwidth]{figure/introduction}}
%%\vspace{-10pt}
%\end{figure}

% 1. 基于IB理论，我们可以通过信息论的角度来分析这个trade-off
% 2. 有趣的是，在IB理论里也存在着一个trade-off，即information-bottleneck trade-off，IB trade-off 描述的是，network learning can be interpreted by finding the optimal trade-off between input information compression and label-related information enhancement.
% 3. 这个trade-off intrigues us去解释我们发现的discriminability-transferability trade-off；
% 4. 通过可视化IB中提出的mutual information dynamics in the information plane. 我们揭露了一个over compression的现象，即在input information compression的过程中，丢弃了和downstream task相关的信息，这些信息不足，导致了下游transfer的性能较差；

%Thanks to the insightful information-bottleneck~(IB) theory, we are allowed to investigate the discriminability-transferability trade-off from the information perspective. 
Interestingly, we notice a concept \textit{Information-Bottleneck~(IB) trade-off} in the IB theory~\cite{ib3,ib2,ib1}, \ie, the network learning can be interpreted by finding the optimal trade-off between input information compression and label-related information enhancement. This IB trade-off intrigues us to explain our observed trade-off between discriminability and transferability.
Following IB, we visualize mutual information dynamics~\cite{mine} in the information plane~\cite{ib2}, and reveal the \textit{over-compression} phenomenon, \ie, prolonged input information compression leads to inadequate information on downstream tasks and thus poor transferability. 

%Counteracting the over representation compression provides a direction to make the transferability compatible with discriminability.

%observe that the improvement in discriminability is attributed to the representation compression phase, where the mutual information between inputs and representations gradually decreases with training progressing.

% 1. 基于IB的解释，同时也提供给了我们一个思路去，让特征同时具有两种属性。我们认为用CE loss驱动模型学习的同时，去counteract the over compression是一个好的思路；
% 2. 在设计方法之前，我们假设了迁移性的解释，
% 2. To this end, 我们explore了InfoNCE loss和counteract over compression的关系，我们设计了一种基于InfoNCE loss的思路，来达到D和T兼容的目的；
% 3. 基于这个思路，我们进一步提出了一种two-stage的学习方法，通过develop两个stage之间的infoNCE loss，我们force第二个stage去counteract over compression；
% 4. 

The above IB-based perspective also provides us an idea to make the discriminability and transferability compatible, \ie, simultaneously training the model with a loss for the specific task and another loss to \textit{counteract the over-compression}. To support this standpoint, we establish the correlation between counteracting over-compression and improving transferability via a principle components analysis~(PCA) perspective. We further explore and provide a solution based on the InfoNCE~\cite{cpc,cmc,infomax} loss to counteract the over-compression. 
Concretely, we present a two-stage learning framework, namely Contrastive Temporal Coding~(CTC). The learning process of CTC consists of two stages, \ie, the information aggregation and revitalization stages. In the first stage, a classification model is optimized and the last epoch model is stored as the information bank, which aggregates informative representations. In the second stage, we introduce an InfoNCE loss between the current model and the information bank, counteracting the over-compression.  As shown in the right part of Figure~\ref{fig:introduction}, our proposed method successfully alleviates the incompatibility, achieving high-quality representations in supervised learning.

\section{Related Work}
% augmix, moco, comparison?
% gradaug, supcon
%\TODO{Comparison to supervised contrastive learning, shorter training epochs, better robustness?, and better transfer performance, different motivation.}

\noindent\textbf{Information-Bottleneck Theory.}
The information-bottleneck~(IB) theory~\cite{ib3,ib2,ib1} provides an information-theoretic principle for encoding the input data into a compressed representation. The theory is based on measuring the mutual information between the input/label variable and the representation variable. It is demonstrated that the representation learning of deep neural networks undergoes two phases, \ie, the empirical error minimization and representation compression phases. In the first phase, the mutual information on the label variable is rapidly increased. When it comes to the compression phase, most of the optimization epochs are spent on decreasing the mutual information on the input variable. 
From the perspective of IB theory, we analyze the incompatibility between the transferability and the discriminability of learned representations in the common supervised learning, which is mostly ignored by the community. We conjecture and prove that the drop of transferability owes to the prolonged compression phase, \ie, the information relevant to downstream tasks is overly discarded for learning discriminative representations. 
%Inspired by the IB theory, we propose a method to maintain the transferability in the supervised learning based on mutual information maximization.

\noindent\textbf{Discriminability vs. Transferability.}
The transferability of deep representations has been studied from various perspectives~\cite{transfer1,transfer2,transfer3,transfer4,transfer5,transfer6,transfer7,transfer8}. And many previous works also noticed the correlation between discrimiability and transferability in the domain adaptation area~\cite{transfer7,transfer8}. In this work, we venture to study the correlation between discriminability and transferability with from an information-theoretic point of view. It is worth noting that~\cite{transfer1} also noticed that better ImageNet classification results~(obtained from better loss functions) could lead to worse transfer learning performances, strongly supporting our work.

\noindent\textbf{Contrastive Learning.}
The main idea of contrastive learning is maximizing the similarity between samples from the same category/view while minimizing the similarity between samples from different categories/views. In recent years, contrastive learning has shown great potentials in self-supervised and unsupervised learning~\cite{simclr,moco,supcon,cpc,contrastiveapp2,cmc,infomin,instancedis,simclr,simclrv2,xiao2017joint,improving} but is not widely studied in the supervised learning setting~\cite{supcon,Shao_2021_WACV}. Different from these previous works, our method contrasts the representations from the current training against those from the previous training. 
The contrastive learning objective has been proven to be the lower bound of the mutual information between the two views~\cite{cpc,cmc,mi_infonce}. In this work, we utilize this property for optimizing the mutual information between deep representations from different training stages. 
%Our experiments demonstrate that our method achieves better transferability than the SupCon~\cite{supcon} with relatively shorter training epochs.

%\subsection{Data Augmentations}
% mixup, cutout, cutmix, auto augmentations, dropout
% augmix, supcon
%The data augmentation technique increases training data by applying linear or non-linear transformations to the original data.
%Several methods have been developed for data augmentation, \eg, CutOut~\cite{cutout}, CutMix~\cite{cutmix} and MixUp~\cite{mixup}. Other than manually designed methods, the AutoAugment method~\cite{autoaug,dada,fastaa} searched from a large space with reinforcement learning. 
%In this work, we find that strong augmentations could help the model maintain a high transferability during training. Combining our method with strong augmentations contributes to better transferability, demonstrating that our CTC is scalable and orthogonal to techniques for improving the transferability.

%\newpage

\section{Discriminability vs. Transferability}
\label{sec:analysis}
In this section, we first reveal the incompatibility between \textit{discriminability} and \textit{transferability} properties of deep representations. Then, we give theoretical and empirical explanations from the information-theoretic perspective. 

\subsection{Revealing the Incompatibility}
\label{subsec:incompatibility}
We optimize a classification model with the cross-entropy loss on CIFAR-100~\cite{cifar} and evaluate these properties\footnote{Representations refer to the outputs of the backbone, which are processed with a global average pooling in popular models~\cite{resnet}. }:

%%\vspace{5pt}
\noindent \textbf{Discriminability.}
Intuitively, the discriminability of representations is reflected by the classification accuracy. However, the high accuracy only indicates separable representations since a sample can be correctly classified but near located to the decision boundary. Previous work~\cite{triplet,centerloss} pointed out that the discriminability can be better revealed by nearest neighbor search~(NNS) algorithms. Besides, a subtle but essential component in a classification model should be considered, \ie, the classifier. Recent works~\cite{bbn,decouple} revealed that the performance of a classification model is closely related to the quality of its classifier. To precisely quantify the discriminability, we propose to evaluate retrieval and clustering performances of representations, which are built upon classifier-irrelevant NNS algorithms. 

To measure discriminability, we evaluate representations with a typical network architecture ResNet18~\cite{resnet}. Representations from each training epoch are extracted and assessed by retrieval and clustering tasks. Recall@$1$~($R$@$1$) is the evaluation metric for retrieval tasks. Normalized Mutual Information~(NMI) is reported to assess clustering performances. 
Metrics are calculated only by the test set to avoid problems that can be caused by over-fitting the training set.

%Hence, to precisely quantify the discriminability, we propose to evaluate retrieval and clustering performances of deep representations, which can be classifier-irrelevant tasks. More discriminative representations can undoubtedly yield better nearest neighbor searching and clustering results. 

%Within expectations, better classification accuracies are gained with more discriminative representations. 

%\subsection{Robustness}
%
%For evaluating the robustness, corrupted and perturbed images are used to test the mean corrupted error~(mCE)\cite{mce} metric of the deep representations. Generally speaking, $15$ kinds of distortions are introduced into images of the test dataset, and models with lower mCE are regarded as more robust ones. 
%Following the setting in~\cite{mce}, ResNet-18 and WRN-28-4 are trained on the CIFAR-100 dataset while tested on the corrupted version datasets CIFAR-100-C. 
%
%Results are reported in Figure~\ref{fig:temporal_analysis}. As observed, the prediction errors on corrupted images can greatly decrease with the training process on both networks. On the seen domain, the generalizability of deep representations is progressively enhanced. It is surprised to observe that models can still make satisfactory predictions on corrupted images in the last epochs, indicating that optimizing the model with relatively small learning rates does not hurt the robustness.
%The information discarding behavior is beneficial for the robustness of deep neural networks. 
%In other words, //

%%\vspace{5pt}
\noindent \textbf{Transferability.}
For measuring the transferability, it is reasonable to transfer learned knowledge to out-of-sample datasets. Given a deep classification model learned on a source dataset, we freeze its backbone network and re-train a classifier on top of the last feature layer on unseen target datasets. Corresponding classification accuracies reveal the transferability.

To measure the transferability, CIFAR-100 is the source dataset, and target datasets are CINIC10~\cite{cinic10} and STL-10~\cite{stl10}, respectively. Following the above experimental settings, ResNet-18 model is utilized.

\begin{figure*}[h]
\begin{center}
\includegraphics[width=1.0\linewidth]{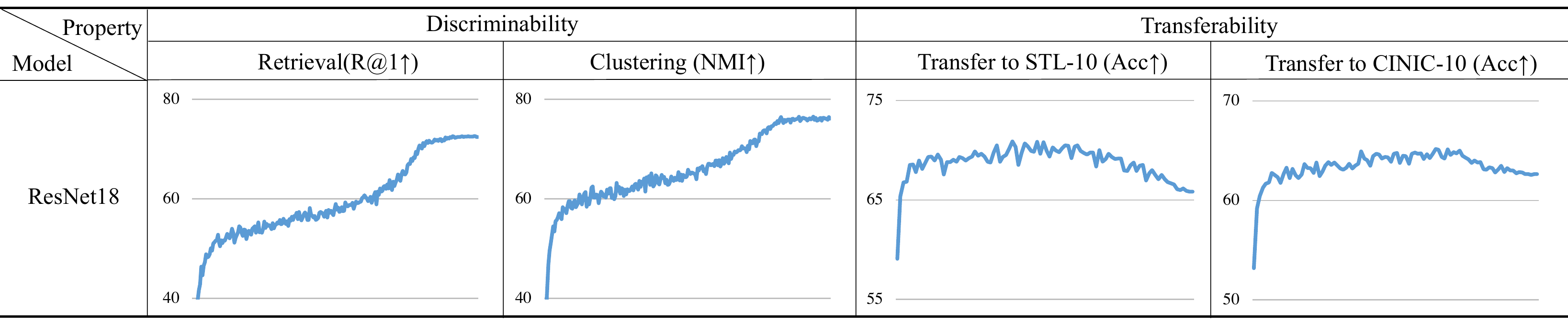}
\end{center}
% %\vspace{-15pt}
\caption{Temporal analyses of representations in the vanilla training process. In each subfigure, the X-axis is the training epoch, and Y-axis is the evaluated metric~(\eg, R@1, NMI, and Top-1 accuracy). ``$\uparrow$'' denotes ``the higher the better''. In the entire training, it can be observed that the discriminability keeps becoming better while the transferability first goes to a peak and then intensely drops in the later training. In a typical supervised training process, discriminability and transferability are incompatible.}
% %\vspace{-15pt}
\label{fig:temporal_analysis}
\end{figure*}

%%\vspace{5pt}
\noindent \textbf{Trade-off between discriminability and transferability. }
\label{sec:discussion} 
As illustrated in Figure~\ref{fig:temporal_analysis}, it can be observed that both $R$@$1$ and NMI have been improved with the training process. However, the continual training in later epochs can significantly hurt the transferability. In all the above experiments, models with the best transferability are mostly located in the middle training stage, rather than the later stage, where the best discriminability is scored.
Conclusively, with the training progressing, representations become increasingly discriminative while the transferability remains uncertain, and, more importantly, these two properties could be incompatible in the later training stage.

\subsection{Connection to Information-Bottleneck Trade-Off}
\label{subsec:info_dynamic}

The trade-off between discriminability and transferability in Sec.~\ref{subsec:incompatibility} motivates us to explain it with the well-known Information-Bottleneck~(IB) trade-off~\cite{ib1,ib2,ib3}.

%%\vspace{5pt}
\noindent \textbf{Recap of the IB theory. }
 The IB theory explains the learning of deep neural networks~(DNNs) by the Information-Bottleneck trade-off, \ie, the network learning is interpreted by finding the optimal trade-off between input information compression and label-related information enhancement. 

An essential viewpoint of IB theory is that, except for the input variable $X$ and label variable $Y$, the hidden representation layer $T$ is regarded as a variable. Under these assumptions, the mutual information~(MI) between $X$~(or $Y$) and $T$ is used to describe the trade-off between input information compression and label-related information enhancement. $I(X;T)$ denotes the MI between inputs $X$ and representations $T$, and $I(T;Y)$ represents the MI between representations $T$ and labels~$Y$. 
Based on MI, the empirical error minimization~(ERM) and representation compression phases are defined. The fast ERM phase rapidly increases $I(T;Y)$, and subsequently, the much longer representation compression phase results in the decrease of $I(X;T)$. These two phases indicate that network first rapidly memorizes label-related information, and then compresses input information for finding an optimal trade-off. 

% 导致网络只抽取和source有关对信息，过于拟合在source上，导致在target上不太好
% discriminability
%%\vspace{5pt}
\noindent \textbf{IB trade-off meets discriminability-transferability trade-off. }
In the IB trade-off, we suppose that input information could be \textit{overly compressed} on the source dataset. Due to the network focuses on enhancing label-related information, the task-irrelevant information could be discarded. Thus, insufficient information on target datasets brings about the unsatisfactory transferability.
In other words, the over-compression results in the aforementioned empirical observation of discriminability-transferability trade-off. To prove this standpoint, we follow IB to calculate MI dynamics on both source and target datasets, capturing the correlation between the information-bottleneck trade-off and discriminability-transferability trade-off~\footnote{We use the Mutual Information Neural Estimation~(MINE)~\cite{mine} method to calculate the mutual information between continuous variables.}.

%The input information is measured by the mutual information between the input variable X and the latent representations T.
%A DNN is regarded a Markov chain of representation~(hidden) layers learned by minimizing the empirical error over the weights of the network. 
%Following IB theory~\cite{ib3,ib2,ib1}, we define the input and output variables as X and Y.
%networks tend to spend most of training epochs to compress representations.

\setlength{\intextsep}{0pt}%
\setlength{\columnsep}{5pt}%
\begin{wrapfigure}{r}{0.5\linewidth} %[t!]
%\vspace{-10pt}
\begin{center}
\includegraphics[width=1.0\linewidth]{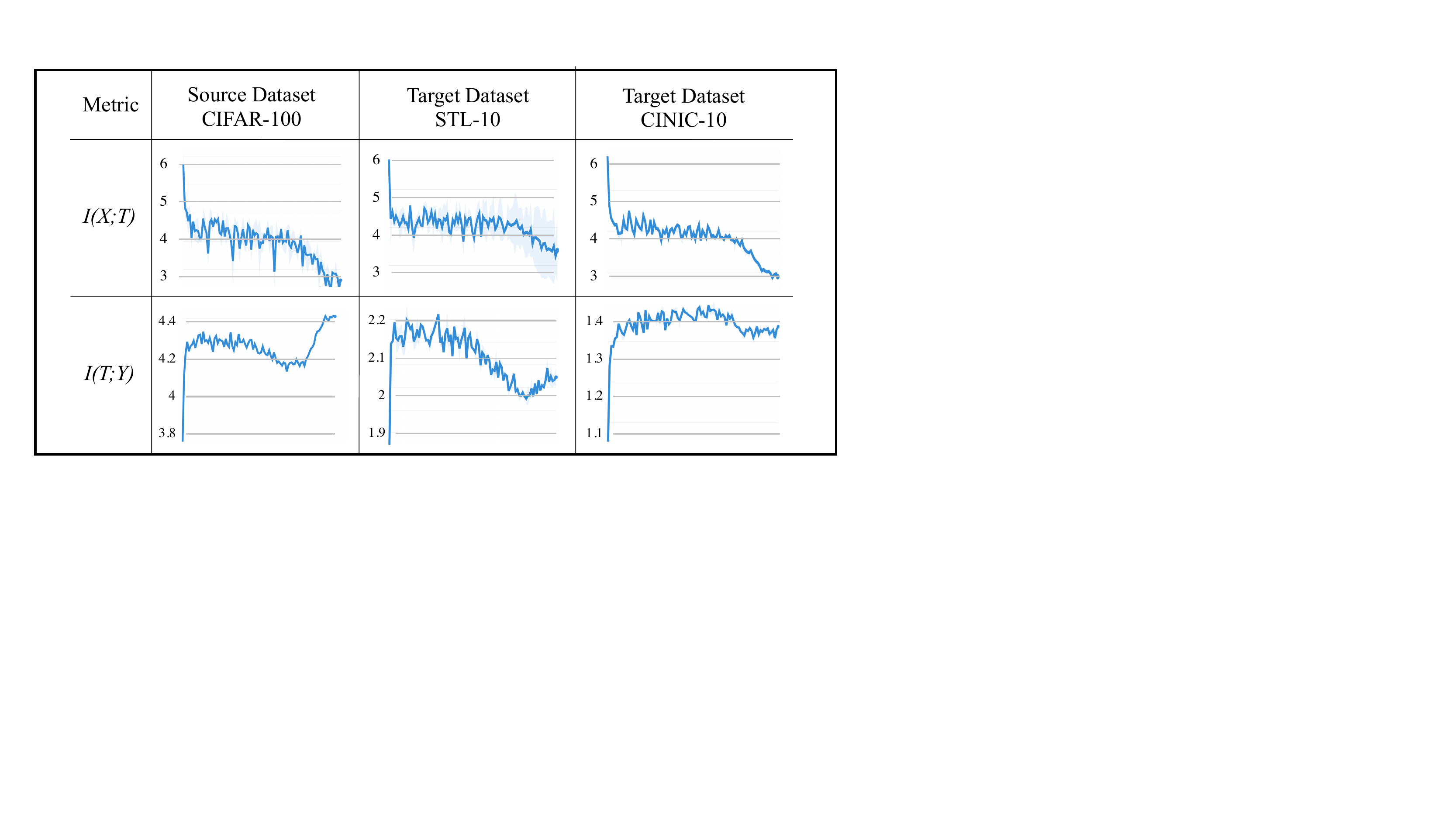}
\end{center}
%\vspace{-25pt}
\caption{Mutual information on both source and target datasets.}
%\vspace{-10pt}
\label{fig:mi_statistics}
\end{wrapfigure}

\textit{(1)~MIs on the source dataset.} As shown in Figure~\ref{fig:mi_statistics}, expectedly, $I(T;Y)$ on the source dataset~(CIFAR-100) rapidly increases in initial training, corresponding to the fast ERM phase. Then, $I(X;T)$ slowly decreases with the training progressing, matching the representation compression phase. Associated with results in Figure~\ref{fig:temporal_analysis}, the discriminability greatly benefits from the representation compression phase. 

\textit{(2)~MIs on the target dataset.} Both $I(X;T)$ and $I(T;Y)$ on target datasets are illustrated in Figure~\ref{fig:mi_statistics}. With representation compression on the source dataset, $I(T;Y)$ on both STL-10 and CINIC-10 climbs to a peak and then gradually decreases, consistently acting like the transferring performances in Figure~\ref{fig:temporal_analysis}. Meanwhile, $I(X;T)$ on target datasets also gradually decrease with the training progressing, indicating that the information relevant to target datasets is also discarded due to the long representation compression phase.

%%\vspace{5pt}
\noindent \textbf{Over-compression leads to the incompatibility. }
The above results reveal the \textit{over-compression} issue, which degrades both $I(X;T)$ and $I(T;Y)$ on target datasets in the later training. Driven by the IB trade-off, the transferability can be sacrificed by the supervised leaning to overly compress input-related information for efficient representations. Over-compression results in insufficient information related to target datasets, unsatisfactory transferring performances, and the incompatibility.  However, we suppose that high-quality representations are expected to possess both properties, arising our next explorations. 
% 为了D去牺牲了T，CE会追求极致的D --- zbr

%It also can increase the $I(X;T)$ on the target dataset by preserving more .

\section{Method}
%We notice that self-supervised learning has shown great potential in transfer learning tasks, while linear probing results of these methods still cannot catch up with supervised learning results. It is similar to supervised learning that the transferability and discriminability also cannot be compatible in self-supervised learning. How to make these properties compatible is well worth studying.

% In this section, we first figure out how to alleviate the observed incompatibility, and then propose a method to achieve this goal.

%4.1.1的逻辑
%1. 解决我们在第三章里发现的incompatibility是一个合理的方向；
%2. 解决它的本质，是对抗I(X;T)下降，即缓解compression；
%3. 根据[CKIM paper]，我们可以直接来优化H(T)做到这件事情；

\subsection{Alleviating the Incompatibility}

\noindent \textbf{Understanding the transferability. }
Following~\cite{dan}, we hypothesize that the transferability between a source dataset and a target dataset relies on the model learns important representation patterns shared by source and target datasets. Thus, the decrease of transferability is attributed to the information of representations which related to the target dataset is overly compressed.

%%\vspace{5pt}
\noindent \textbf{Counteracting over-compression improves transferability. }
To achieve this goal, an intuitive solution is to counteract the decrease of $I(X;T)$ on the source dataset. To support our motivation, we give an explanation from the Principal Components Analysis~(PCA) perspective.

Given an example where the model is linear, maximizing $I(X;T)$ is equivalent to solving a PCA on the source dataset\footnote{Proofs are attached in the appendix A.1.}. Solving PCA helps the model capture the most representative representations on the source dataset. If the target dataset shares important patterns with the source dataset, maximizing $I(X;T)$ on the source dataset ensures the most important representation patterns are captured, further improving transferring performances. Since calculating the mutual information for continuous variables is challenging, we introduce an approximation in the next. The definition of $I(X;T)$ is as followed:
\begin{small}
$$
 I(X;T) = H(T) - H(T|X),  
$$
\end{small}
where $H(T)$ is the entropy of the latent representation $T$ and $H(T|X)$ is the conditional entropy. Since the support set of $X$ contains tremendous natural images from the $P(X,Y)$ and the neural network is deterministic, the conditional entropy equals 0~\cite{cikm}. Thus, the above equation can be re-written as:
\begin{small}
$$
 I(X;T) = H(T),  
$$
\end{small}
% increasing $H(T)$ or decreasing $H(T,Y)$ are feasible directions. However, labels $Y$ of the target dataset are not accessible when training on the source dataset. 
% 我们可以优化H（T），达到同时提升I（X；T）和I（Y；T）的目的
which demonstrates that counteracting the decreasing of $H(T)$ on the source dataset could be a potential way to alleviate over-compression. 
%Given T is parameterized by the neural network and not related to target data, we are allowed to counteract the decrease of $H(T)$ on the source dataset by .

%Due to we cannot access $T$ on the target dataset, the left choice is counteracting the decreasing of $H(T)$ on the source dataset. We suppose that $H(T)$ is generalizable, and both $I(X;T)$ and $I(T;Y)$ on the target dataset will be increased~(proven in later experiments).
%learning informative representations on the source dataset will lead to better transferability. 

%
%4.1.2的逻辑
%1. 有paper指出infoNCE可以做到I(T;T)的优化；
%2. 那我们可以先获得一个T，当作lower bound，再去优化另一个T；

%\noindent \textbf{Self-supervised learning and information theory. }  

%%\vspace{5pt}
\noindent \textbf{Counteracting the decrease of $\mathit{H(T)}$. }
In the following, we demonstrate how to counteract the decrease of $H(T)$.
It is proven in~\cite{infomax,cpc,cmc} that minimizing the InfoNCE loss maximizes a lower bound on mutual information. Given two representation variables $T_1$ and $T_2$, the relation between InfoNCE loss and the mutual information $I(T_1;T_2)$ can be derived as:
\begin{small}
\begin{equation}
%    I(X, T) \geq log(N) - \sum_{i=1}^{N} \mathcal{L}_{\text{IAS}}.
    I(T_1; T_2) \geq \text{log}(N) - \mathcal{L}_{\text{InfoNCE}},
\label{eq:1}
\end{equation}
\end{small}
where $N$ is normally the number of samples in a training set.
As demonstrated, minimizing the InfoNCE loss would increase the lower bound of $I(T_1;T_2)$. Combining Eqn.(\ref{eq:1}) with the definition of the mutual information:
\begin{small}
\begin{equation}
\begin{aligned}
I(T_1;T_2) & = H(T_1) + H(T_2) - H(T_1,T_2) \\ 
		   & \leq  H(T_1) + H(T_2) - \text{max}(H(T_1), H(T_2)), \\ 
		   & = \text{min}(H(T_1), H(T_2)), \nonumber
\label{eq:mutual_info}
\end{aligned}
\end{equation}
\end{small}
the Eqn.(\ref{eq:1}) could be re-written as:
%\begin{align}
%\text{log}(N) - \mathcal{L}_{\text{InfoNCE}} \leq
%\begin{cases}
%H(T_1),  & \text{if } H(T_1) < H(T_2) \\
%H(T_2), & \text{if } H(T_1) \geq H(T_2).
%\end{cases}
%\label{eq:mutual_lower_bound}
%\end{align}
\begin{small}
\begin{align}
\text{log}(N) - \mathcal{L}_{\text{InfoNCE}} \leq \text{min}(H(T_1), H(T_2)).
\label{eq:mutual_lower_bound}
\end{align}
\end{small}
Eqn.(\ref{eq:mutual_lower_bound}) suggests that minimizing InfoNCE loss improves the lower bound of $\text{min}(H(T_1), H(T_2))$, which simultaneously improves lower bounds of $H(T_1)$ and $H(T_2)$. 
Therefore, we could select the representation $T_1$ from early training and \textit{fix it as constant}, regard the representation in later training as $T_2$, and develop an InfoNCE loss between the constant representation $T_1$ and the later representation $T_2$, for counteracting the decrease of $H(T_2)$. Inspired by the well-known ``memory bank'' concept, we name the constant representation $T_1$ as the \textit{information bank}. 

Concretely, in Eqn.(\ref{eq:mutual_lower_bound}), if $H(T_2) \leq H(T_1)$, the InfoNCE loss would improve the lower bound of $H(T_2)$. If $H(T_2) > H(T_1)$, the objective of counteracting has been reached. In this manner, the loss $\mathcal{L}_{\text{InfoNCE}}$ ensures the representation of later training has a relatively large $H(T)$.

\subsection{Contrastive Temporal Coding}

%既然D和T之间是存在不兼容的，为了缓解两者的冲突，我们用一种progressive的方法来训，我们希望先得到一个好的D，再去提高特征的T；

Inspired by the above explorations, we propose a two-stage learning framework to alleviate the incompatibility between discriminability and transferability. In the first training stage, named information aggregation stage, the main objective is to obtain the information bank to provide the $H(T_1)$ in Eqn.(\ref{eq:mutual_lower_bound}). In the second training stage, named information revitalization stage, the main objective is to further counteract the decrease of $H(T_2)$ via Eqn.(\ref{eq:mutual_lower_bound}).
%To alleviate the incompatibility, we propose a progressive two-stage framework to train the model, and two stages emphasize discriminability and transferability, respectively.

%, maximize corresponding mutual information and preserve the transferability property of deep representations.

%2. 我的方法就是两阶段的，第一阶段正常训，第二阶段在第一阶段中挑一个T1，第二阶段中去保住T，但是如果第一阶正常训的话T掉的太快了，很难找到一个D不错T也不错的东西，所以我们直接在一阶段也用infoNCE，保证最后一个epoch就可以用，解决选bank的问题

%\begin{table}[t]
%\centering
%\begin{scriptsize}
%\begin{tabular}{l|c}
%\Xhline{2\arrayrulewidth}
%method                        & CIFAR-100 top-1~(\%)        \\ \hline
%ResNet18+CosLr          & 79.3$\pm$0.2                    \\
%ResNet18+CTC(Ours)      & \textbf{80.1}$\pm$0.3  \\ 
% \Xhline{2\arrayrulewidth}
%\end{tabular}
%\end{scriptsize}
%%%\vspace{-10pt}
%\caption{Top-1 accuracies~(\%) on CIFAR-100 of baseline and CTC. ``CosLr'' denotes training the model with the cosine learning rate scheduler. Results are averaged on five runs. }

%%\vspace{5pt}
\noindent \textbf{Information aggregation stage~(IAS). }
In this stage, a classification model is trained with a vanilla cross-entropy~(CE) loss $\mathcal{L}_{\text{CE}}$ as the main loss function for learning $T_1$. Eqn.(\ref{eq:mutual_lower_bound}) indicates that, if $H(T_2) \leq H(T_1)$, the InfoNCE loss will improve the lower bound of $H(T_2)$. Thus, it is reasonable to choose the $T_1$ with a relatively large $H(T_1)$ in this stage. However, experiments in Sec.~\ref{sec:analysis} demonstrate that, only with the CE loss, the information compression is not controllable and predictable, and the last epoch model could still be overly compressed.
For practical usage, it is tricky to select a good $T_1$ among all epochs, and we hope the last model is a satisfactory choice. To this end, we introduce an auxiliary InfoNCE loss $\mathcal{L}_{\text{IAS}}$ for desensitizing the model selection.

Let $T_1$ denote the representation variable. Similar to IB theory, we regard a memory bank $V$ as a variable, which keeps representations for each training sample and provide contrastive samples. For $t_1 \in T_1$ and $v \in V$~(which are normalized representations), contrastive representation pairs are composed as $\{ t_1^i, v^j\}_{i,j=1}^{N}$, and $N$ is the number of training samples. Positive pairs are constructed by representations of the identical samples, while negative pairs are composed of different samples. 
%To pull similar samples ``close'' and push dissimilar samples ``away'', a function $h(\cdot, \cdot)$ is learned to discriminate positive and negative pairs. 
%Thus, the NCE-style optimization objective function can be written as:
%\begin{equation}
%	\mathcal{L}_{\text{IAS}} = - \mathop{\mathbb{E}}\limits_{\{z_1^i, x^1, \ldots, x^N\}} \left[ \text{log} \frac{h(z_1^i, x^i)}{\sum_{j=1}^{K} h(z_1^j, x^i)} \right],
%\end{equation}
%Then the mutual information $I(X;T)$ can be maximized with the minimization of $\mathcal{L}_{\text{IAS}}$: 
%\begin{equation}
%%    I(X, T) \geq log(N) - \sum_{i=1}^{N} \mathcal{L}_{\text{IAS}}.
%    I(X, T) \geq log(N) - \mathcal{L}_{\text{IAS}}.
%\end{equation}
%However, this loss function could be unsuitable for direct optimization. Firstly, information in raw images is merely semantic, and high-level representations may not benefit from this information. Secondly, the discriminating function could be extremely large for raw images of large resolutions. Instead, maximizing the MI between the high-level representations is a better choice. 
Therefore, the loss function of $t_1^i$ can be re-written as:
%\begin{equation}
%	\mathcal{L}_{\text{IAS}} = - \mathop{\mathbb{E}}\limits_{\{t_1^i, v^1, \ldots, v^N\}} \left[ \text{log} \frac{h(t_1^i, v^i)}{\sum_{j=1}^{K} h(t_1^j, v^i)} \right],
%\end{equation}
\begin{small}
\begin{equation}
	\mathcal{L}_{\text{IAS}} = -  \text{log} \frac{\text{exp}(t_1^i \cdot v^{i})}{\sum_{j=1}^{N} \text{exp}(t_1^i \cdot v^j)},
\end{equation}
\end{small}
where $v^{i}$ is the positive ``key''. When the learning process ends, the model of the latest epoch is saved as the information bank $\hat{T}_1$. 
%$N$ denotes the number of training samples.

%Positive pairs are constructed by samples from the same category, while samples in a negative pair are from different categories. To pull similar samples ``close'' and push dissimilar samples ``away'', a function $h(\cdot, \cdot)$ is learned to discriminate positive and negative pairs.

%and $\{\hat{t_1}^j\}_{j=1}^{K}$ are $K$ representation samples~(containing one positive sample $\hat{t_1}^k$ and $K-1$ negative samples $\{\hat{t_1}^j | j \neq k\}$)

%%\vspace{5pt}
\noindent \textbf{Information revitalization stage~(IRS). }
Let $T_2$ denote representation variable of the second stage. In this stage, we continue training the model from the end of the former stage, and develop an InfoNCE loss $\mathcal{L}_{\text{IRS}}$ to counteract the decrease of $H(T_2)$.  For $\hat{t}_1 \in \hat{T}_1$ and $t_2 \in T_2$, contrastive representation pairs are composed as $\{\hat{t}_1^i, t_2^j\}_{i,j=1}^{N}$,
%Following popular frameworks for contrastive learning, given a representation $t_2^i$ from the current epoch, we construct a subset of contrastive pairs $\{\hat{t_1}^j, t_2^i\}_{j=1}^{K}$ to optimize the function, 
where $t_2$ is from the current epoch and $\hat{t}_1$ is from the previously saved information bank. 
The optimization objective function of $t_2^{j}$ can be written as:
%\begin{equation}
%	\mathcal{L}_{\text{IRS}} = - \mathop{\mathbb{E}}\limits_{\{t_2^k, \hat{t}_1^1, \ldots, \hat{t}_1^K\}} \left[ \text{log} \frac{h(\hat{t}_1^k, t_2^k)}{\sum_{j=1}^{K} h(\hat{t}_1^j, t_2^k)} \right].
%\end{equation}
\begin{small}
\begin{equation}
	\mathcal{L}_{\text{IRS}} = -  \text{log} \frac{\text{exp}(t_2^j \cdot \hat{t}_1^{j})}{\sum_{k=1}^{N} \text{exp}(t_2^j \cdot \hat{t}_1^{k})},
\end{equation}
\end{small}
where $\hat{t}_1^{j}$ is the positive ``key''.
%where $K$ is the number of representation samples from the information bank model and $h(\cdot, \cdot)$ is the aforementioned discriminating function. 
%According to~\cite{cpc,cmc}, by minimizing the objective function $\mathcal{L}_{\text{CTC}}$, the mutual information of representations from both models $\mathit{I}(z_1; z_2)$ is maximized, which can be formulated as:
%\begin{equation}
%    I(z_1; z_2)\geq log(K) - \mathcal{L}_{\text{CTC}}.
%\end{equation}
Consequently, the optimization objective can implicitly counteract the decrease of $H(T_2)$ and over-compression, promoting the transferability of learned representations. 
%A formal proof is given in our supplementary materials.

%\paragraph{\noindent \textbf{Learning Framework.}} As illustrated in Figure~\ref{fig:framework}, the learning of the information aggregation stage is singly driven by a cross-entropy loss function $\mathcal{L}_{\text{CE}}$. At the end of this stage, an imperfect model is saved. In the information preservation stage, the imperfect model is frozen and only supplies deep representations for the temporal contrastive learning. The learning of the current model is simultaneously motivated by a contrastive learning loss $\mathcal{L}_{\text{CTC}}$ and a cross-entropy loss $\mathcal{L}_{\text{CE}}$. The total loss can be written as:
%%\vspace{5pt}
\noindent \textbf{Learning framework.}
The learning of the information aggregation stage is driven by a cross-entropy loss function $\mathcal{L}_{\text{CE}}$ and a contrastive loss $\mathcal{L}_{\text{IAS}}$ as:
\begin{small}
\begin{equation}
	\mathcal{L}_{\text{stage 1}} = \alpha  \mathcal{L}_\text{IAS} + \mathcal{L}_\text{CE}.
\end{equation}
\end{small}
At the end of this stage, the information bank model is saved. In the information revitalization stage, the learning of the current model is simultaneously motivated by a contrastive learning loss $\mathcal{L}_{\text{IRS}}$ and a cross-entropy loss $\mathcal{L}_{\text{CE}}$. The $\mathcal{L}_{\text{IRS}}$ is calculated with representations from both the current model and the information bank. The loss can be written as:
\begin{small}
\begin{equation}
	\mathcal{L}_{\text{stage 2}} = \beta  \mathcal{L}_\text{IRS} + \mathcal{L}_\text{CE}.
\end{equation}
\end{small}
$\alpha$ and $\beta$ are weighting hyper-parameters, which respectively emphasize the importance of two properties. The general framework could be similar to the well-known knowledge distillation process~\cite{kd,ban,dkd}; however, learning the first stage without the $\mathcal{L}_\text{IAS}$ loss function~(which corresponds to the teacher model training) still leads to poor transferability in $T_1$. Consequently, developing the $\mathcal{L}_\text{IPS}$ loss with such $T_1$ will limit the increasing of $T_2$.

\subsection{Discussions on Self-Supervised Learning~(SSL)}

%According to Eq.~\ref{eq:mutual_lower_bound} and discussions in Section~\ref{subsec:info_dynamic}, the good transferability of self-supervised learning could result from the learning target, which alleviates the over representation compression by counteracting the decrease of representation entropy.

We notice that SSL has shown great potential in transfer learning tasks~\cite{instancedis,moco,mocov2}, while linear probing results of these methods still cannot catch up with supervised learning results. It is similar to supervised learning that the transferability and discriminability also cannot be compatible. However, the superiority of SSL in improving transferability drives us to explore the underlying working mechanism.

\noindent \textbf{On transferability. } Mainstream SSL methods depend on InfoNCE loss, and we attempt to formulate its learning target from the information view. In representative works, a dictionary~(memory bank~\cite{instancedis} or momentum encoder~\cite{moco,mocov2}) is typically used for providing negative samples for a trainable encoder. We denote the dictionary representation as $T_1$, and the trainable encoder representation as $T_2$.
Given that $T_1$ comes from the dictionary, the $H(T_1)$ could be regarded as a constant before being updated. Minimizing the contrastive loss also counteracts the decrease of $H(T_2)$. Thus, the good transferability of SSL methods could also result from the learning target for informative representations. 

\noindent \textbf{On discriminability. } Mainstream SSL methods can be interpreted by a ($K$$+$$1$)-way softmax-based classification task, where $K$ is the number of negative samples~\cite{moco}. Intuitively, due to $K$ is a large number, the ($K$$+$$1$)-way classification task is more challenging than the conventional supervised image classification.
Thus, we conjecture that SSL compresses input information slowly and is inferior to enhance label-related information. We also notice in~\cite{mocov2} that longer training will bring in significant discriminability improvement but minor transferability improvement, which suggest that SSL methods could require longer time for compressing input information and enhancing label-related information. 

Later experiments will demonstrate that our CTC outperforms SSL methods on the transferability property.

%\begin{algorithm}
%\caption{Pseudocode of CTC in a PyTorch-like style.}
%\label{algo:CTC}
%\footnotesize
%\begin{alltt}
%\color{OliveGreen}
%# net_1, net_2: networks for two stages
%# t1, t2: numbers of epochs for two stages
%# extract: a function to extract representations
%# alpha, beta: hyper-parameters
%\end{alltt}
%%%%\vspace{-15pt}
%\begin{alltt}
%for _ in t1:
%  for x in loader:
%    logits = net_1.forward(x)
%    loss_ce = CrossEntropyLoss(logits, labels)
%    loss_ce.backward()
%    update(net_1.param)
%    
%net_2.param = net_1.param
%for _ in t2:
%  for x in loader:
%    logits = net_2.forward(x)
%    f_1 = net_1.extract(x).detach()
%    f_2 = net_2.extract(x)
%    loss_CTC = ContrastiveLoss(f_1, f_2)
%    loss_ce = CrossEntropyLoss(logits, labels)
%    loss = alpha * loss_CTC + beta * loss_ce
%    loss.backward()
%    update(net_2.param)
%\end{alltt}
%\end{algorithm}

\section{Experiments}
Experiments and discussions focus on the following two parts:~(1)~Validating our motivation and (2)~Transferring representations learned by our method. 
% Due to the page limit, we omit and attach the dataset introductions and implementation details in our appendix.

\begin{figure*}[t]
\begin{center}
\includegraphics[width=1.0\linewidth]{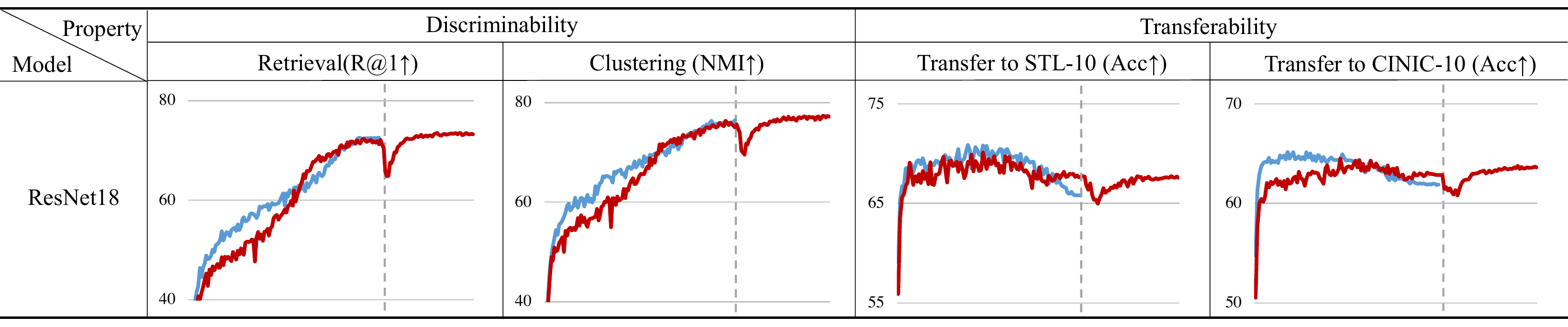}
\end{center}
%\vspace{-15pt}
\caption{Temporal analysis of representations in the \textcolor{lineblue}{\textbf{vanilla}} and \textcolor{linered}{\textbf{CTC}} training processes. Results of the vanilla and CTC are colored in blue and red, respectively. Two stages are divided by grey dotted lines. Compared with the vanilla training, it can be observed that the transferability is greatly preserved by CTC in later epochs. Furthermore, discriminability is improved by preserving the transferability.}
%\vspace{-15pt}
\label{fig:CTC_analysis}
\end{figure*}

\subsection{Motivation Validation and Main Results}
\label{subsec:main_results}

%%\vspace{5pt}
\noindent \textbf{Motivation: counteracting over-compression. } 
%According to Eq.~\ref{eq:mutual_lower_bound} and discussions in Section~\ref{subsec:info_dynamic}, we conjecture that the good transferability of self-supervised learning could result from the learning target, which counteracts the over representation compression by increasing representation entropy. Then, we tailor a two-stage method. In the first training stage, we optimize the model with only a cross-entropy loss and learned representations correspond to $T_1$. Then, in the second stage, we introduce a contrastive loss for maximizing the mutual information between $T_1$ and $T_2$, where $T_2$ denotes representations learned in the second stage. In the following, we prove that optimizing this contrastive loss counteracts representation compression.
In this part, we prove our motivation that the proposed method is able to counteract the over-compression.
\setlength{\intextsep}{5pt}%
\setlength{\columnsep}{5pt}%
\begin{wrapfigure}{r}{0.48\linewidth}
\begin{center}
\includegraphics[width=1.0\linewidth]{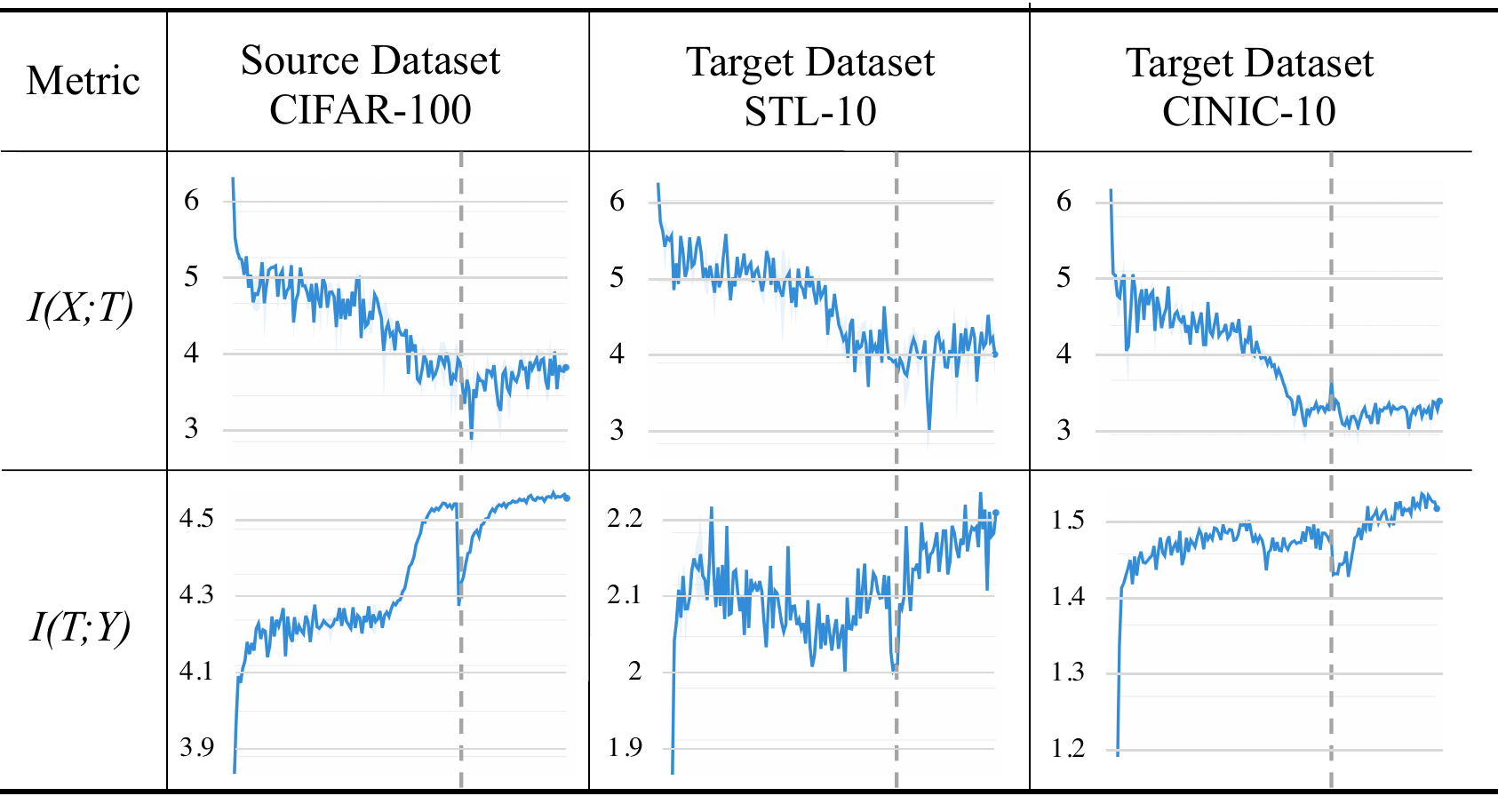}
\end{center}
\caption{Mutual information of CTC on both source and target datasets. For better illustration, two stages are split with grey dotted lines. The left and right parts correspond to information aggregation and revitalization stages, respectively.}
%\vspace{-20pt}
\label{fig:mi_statistics_ctc}
\end{wrapfigure}
Following the experimental settings in Sec.~\ref{sec:analysis}, we report a temporal analysis of our proposed method. Generally, ResNet18~\cite{resnet} is trained with CTC on the CIFAR-100~\cite{cifar} dataset, and representations of each epoch are evaluated and reported afterward. For discriminability, representations of models from each epoch are extracted to perform retrieval and clustering tasks. As to the transferability, we transfer learned representations to STL-10~\cite{stl10} and CINIC-10~\cite{cinic10}.

Evaluations of discriminability and transferability are illustrated in Figure~\ref{fig:CTC_analysis}. 
We divide two stages with a grey dotted line. In the first stage, benefitting from $\mathcal{L}_{\text{IAS}}$, our method~(red lines) achieves better transferring results in the last epoch than the baseline~(blue lines).
The second stage further improves transferring results on target datasets~(STL-10 and CINIC-10) by large margins. Mutual information dynamics are provided in Figure~\ref{fig:mi_statistics_ctc}. Similar to transferring results, the $I(T;Y)$ on target datasets also keeps increasing in the second stage, showing the information relevant to target datasets is revitalized by our CTC. Meanwhile, the $I(X;T)$ on source and target datasets is also non-decreasing in the second stage. The above results jointly demonstrate that the over-compression has been successfully alleviated by our method. Besides, the discriminability is not damaged, proving the incompatibility is also mitigated.

\noindent \textbf{Benchmarking on CIFAR-100 and ImageNet. } To show the direct benefits of counteracting the over-compression, we benchmark our proposed CTC on CIFAR-100~\cite{cifar} and ImageNet~\cite{imagenet}. For CIFAR-100, results are reported in Table~\ref{tab:cifar_tinyimagenet}. Our proposed method consistently outperforms the baseline method. For ImageNet, results are reported in Table~\ref{tab:imagenet} and better results are also achieved compared with the vanilla training results. Counteracting the over-compression does not damage the discriminability and conversely benefits the classification performances. Moreover, our CTC brings in no increase on the number of model parameters and FLOPs.

\begin{table}[h]
%\vspace{-15pt}
\centering
\begin{minipage}{0.47\textwidth}
\centering
\begin{small}
\begin{tabular}{l|c}
\Xhline{2\arrayrulewidth}
method                        & top-1 acc.~(\%)        \\ \hline
Res18+CosLr          & 79.3$\pm$0.2                    \\
Res18+CTC(Ours)      & \textbf{80.1}$\pm$0.3  \\ 
 \Xhline{2\arrayrulewidth}
\end{tabular}
\end{small}
\caption{Top-1 accuracies~(\%) on CIFAR-100 of baseline and CTC~(5 runs).}
\label{tab:cifar_tinyimagenet}
\end{minipage}
\quad
\begin{minipage}{0.47\textwidth}
\centering
\begin{small}
\begin{tabular}{l|c}
\Xhline{2\arrayrulewidth}
method                        & top-1 acc.~(\%)        \\ 
\hline
Res50+CosLr               & 76.1$\pm$0.1                  \\
Res50+CTC(Ours)           & \textbf{76.4}$\pm$0.1                   \\
\Xhline{2\arrayrulewidth}
\end{tabular}
\end{small}
\caption{Top-1 accuracies~(\%) on ImageNet of baseline and CTC~(3 runs). }
\label{tab:imagenet}
\end{minipage}
%\vspace{-25pt}
\end{table}

\begin{figure}[h]
\centering
\begin{minipage}{0.45\textwidth}
\centering
\begin{small}
\begin{tabular}{l|c}
\Xhline{2\arrayrulewidth}
method                        & top-1 acc.~(\%)        \\ \hline
Res18+CosLr          & \textcolor{lineblue}{79.3}$\pm$0.2                    \\
Res18+CTC($\alpha=0.1$)      & 80.1$\pm$0.3  \\ 
Res18+CTC($\alpha=0.5$)      & \textcolor{linered}{79.4}$\pm$0.3  \\ 
 \Xhline{2\arrayrulewidth}
\end{tabular}
\end{small}
\caption{Top-1 accuracies~(\%) on CIFAR-100. We adjust the $\alpha$ from 0.1 to 0.5, and the top-1 accuracy is decreased but still better than vanilla training. Transferring results are in the right figure.}
\label{tab:alpha}
\end{minipage}%
\quad 
\begin{minipage}{0.5\textwidth}
        \centering
        \includegraphics[width=0.8\linewidth]{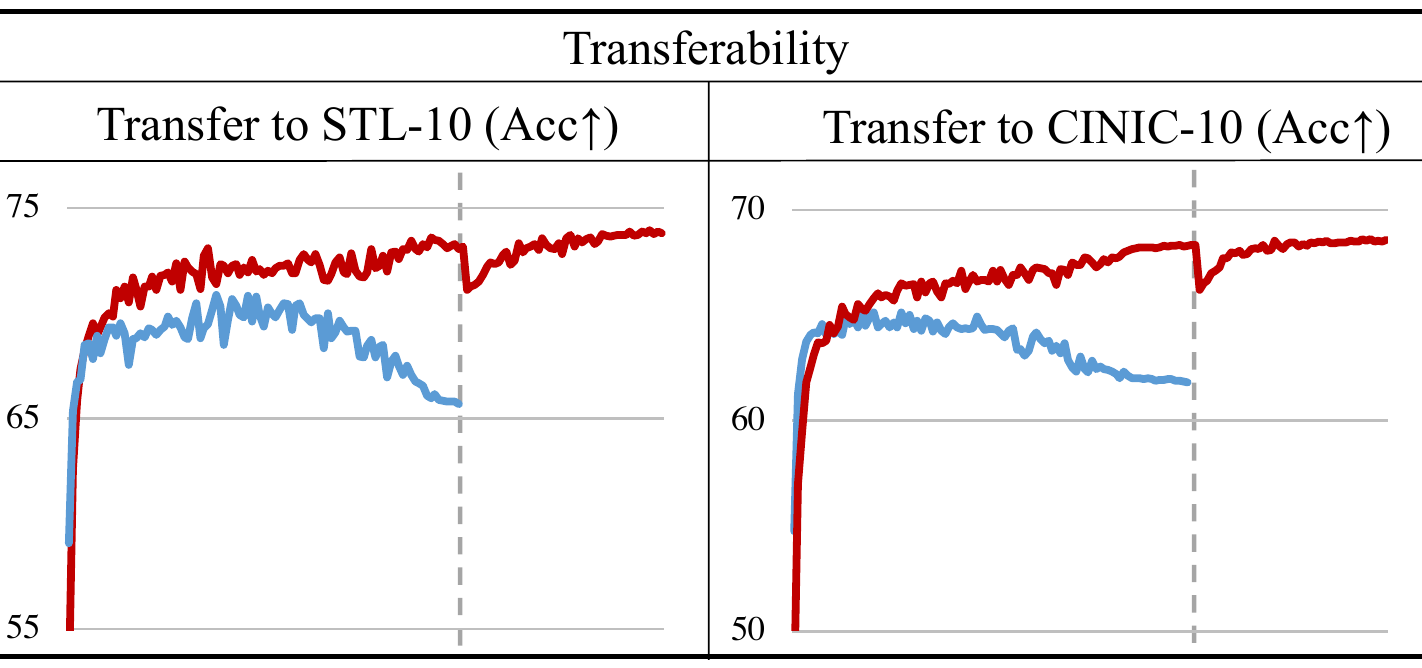}
        %\vspace{-10pt}
        \caption{Setting \textcolor{linered}{\textbf{CTC}}'s $\alpha$ to 0.5~(red) leads to greatly better transferring results than \textcolor{lineblue}{\textbf{vanilla}} training~(blue), showing the over-compression is sufficiently alleviated. }
        \label{fig:alpha}
\end{minipage}
\end{figure}

\noindent \textbf{Boosting transferability with CTC. } To further validate that (1) the correctness of the discriminability-transferability trade-off and (2) our method is able to adjust the trade-off, we conduct experiment on sacrificing the discriminability for boosting the transferability. Since the information aggregation stage of CTC decides the lower bound of $H(T)$, we could adjust the hyper-parameter $\alpha$ for helping the model learn informative representations. Unavoidably, large weight for $\mathcal{L}_{\text{IAS}}$ influences the learning of $\mathcal{L}_{\text{CE}}$. As shown in Table~\ref{tab:alpha} and Figure~\ref{fig:alpha}, increasing $\alpha$ to 0.5 leads to a normal classification accuracy on CIFAR-100, but significantly better transferring results. It also suggests that ending the baseline at an appropriate time is a bad option for enhancing transferability. Moreover, early ending baseline would unavoidably lead to poor discriminability.

\subsection{Towards Better Transferability}
%%%%\vspace{-0.1cm}
In this part, we study (1) how to get better transferability of representations and (2) how to plug in CTC to further boost the transferability, proving its scalability.

\begin{figure*}[t]
\begin{center}
\includegraphics[width=1.0\linewidth]{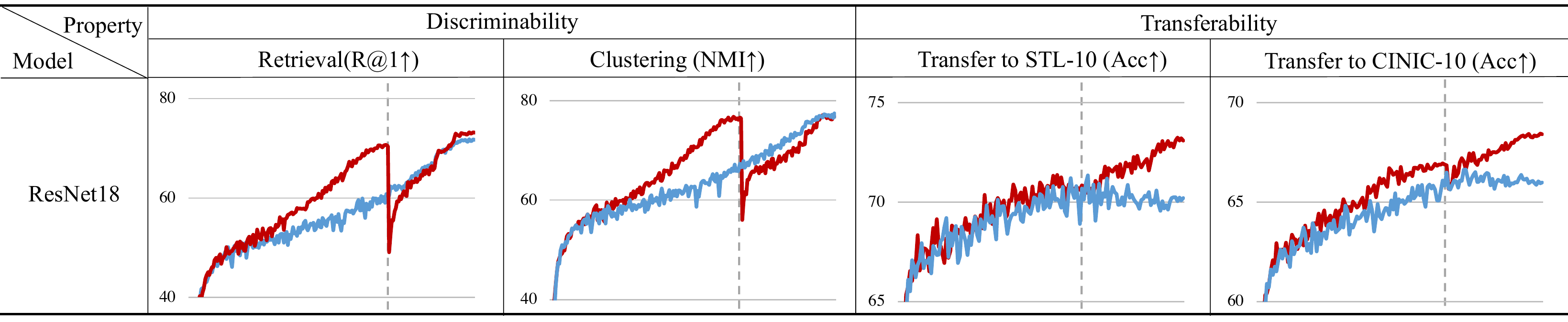}
\end{center}
%\vspace{-20pt}
\caption{Temporal analyses of representations in the \textcolor{lineblue}{\textbf{vanilla}} and \textcolor{linered}{\textbf{CTC}} training \textit{with the AutoAugment}. Results of the vanilla and CTC are colored in blue and red, respectively. Two stages of CTC~(red lines) are separated by the grey dotted line. Compared with the vanilla training~(blue lines), it can be observed that the transferability of CTC keeps climbing in the entire learning process. Different to Figure~\ref{fig:CTC_analysis}, we train the baseline and CTC with the same number of epochs for fair comparisons. }
%%\vspace{-10pt}
\label{fig:CTC_analysis_autoaug}
\end{figure*}

%%\vspace{5pt}
\noindent \textbf{AutoAugment~(AA) contributes to good transferability. } We suppose that strong augmentations could contribute to good transferability, since strong augmentation can increase the difficulty of information compression, and thus lead to higher $H(T)$. Except for normal crop and resize operations, color normalization and translation operations might be more difficult to learn. Thus, on the CIFAR-100 dataset, we first conduct temporal analysis on representations learned with the AA~\cite{autoaug}, and results are blue lines in Figure~\ref{fig:CTC_analysis_autoaug}. Similar to results from Sec.~\ref{sec:analysis}, the discriminability of representations becomes better with the training progressing. However, it is surprised to notice that the transferability of representations learned with the AA does not diminish as intensely as representations learned without AA, especially in later training epochs. It conveys an intuition that strong augmentations could contribute to good transferability, which possibly result from our hypothesis that strong augmentation can increase the difficulty of information compression.

%%\vspace{5pt}
\noindent \textbf{``AA+CTC'' contributes to better transferability.}
%%%\vspace{-0.3cm}
Subsequently, we prove our CTC is scalable and orthogonal to other methods.
We first optimize the model with CTC and AA, and then report the temporal analysis by red lines in Figure~\ref{fig:CTC_analysis_autoaug}. 
With the assistance of CTC, discriminability scores higher within expectations. Moreover, the transferability keeps climbing in the entire learning process, eventually achieving the best transferring classification accuracies which can not be achieved only with AA. It further validates our motivation that the transferability could be promoted by counteracting the decrease of representation entropy. 
For comprehensiveness, we benchmark our CTC with AutoAugment on CIFAR-100 and ImageNet. All experimental settings in this part are the same as those in Sec.~\ref{subsec:main_results}. As reported in Table~\ref{tab:cifar_tinyimagenet_aa} and~\ref{tab:imagenet_aa}, superior classification accuracies than the baseline also meet our expectations.

\begin{table}[t]
\centering
\begin{minipage}{0.47\textwidth}
\centering
\begin{small}
\begin{tabular}{l|c}
\Xhline{2\arrayrulewidth}
method                        & top-1 acc.~(\%)        \\ \hline
Res18+AA+CosLr          & 80.4$\pm$0.3                    \\
Res18+AA+CTC(Ours)      & \textbf{81.2}$\pm$0.2  \\ \Xhline{2\arrayrulewidth}
\end{tabular}
\end{small}
\caption{Top-1 accuracies~(\%) on CIFAR-100 of baseline and CTC with AA.}
\label{tab:cifar_tinyimagenet_aa}
\end{minipage}
\quad
\begin{minipage}{0.47\textwidth}
\centering
\begin{small}
\begin{tabular}{l|c}
\Xhline{2\arrayrulewidth}
method                        & top-1 acc.~(\%)        \\ 
\hline
Res50+AA+CosLr            & 76.8$\pm$0.1                    \\
Res50+AA+CTC(Ours)        & \textbf{77.2}$\pm$0.1                   \\
\Xhline{2\arrayrulewidth}
\end{tabular}
\end{small}
\caption{Top-1 accuracies~(\%) on ImageNet of baseline and CTC with AA. }
\label{tab:imagenet_aa}
\end{minipage}
%\vspace{-20pt}
\end{table}

\subsection{Transferring Representations}
In this part, we present extensive tasks and datasets to which representations learned by our method can be transferred. Since self-supervised learning methods exceed supervised learning in transferring tasks, we also compare with  representative self-supervised learning works~\cite{moco,mocov2,infomin}.
%%%%\vspace{-0.4cm}

\begin{table*}[t]
\centering
\begin{small}
\begin{tabular}{l|c|c|c|c|c|c}
\Xhline{2\arrayrulewidth}
\multirow{2}{*}{pre-training method} & \multicolumn{6}{c}{Performance}                         \\  \cline{2-7}
                        & AP$^{bbox}$ & AP$^{bbox}_{50}$ & AP$^{bbox}_{75}$  & AP$^{mask}$ & AP$^{mask}_{50}$ & AP$^{mask}_{75}$ \\ \hline
Res50 random init. &  30.2       &  48.9              &  32.7               &  28.6         &   46.6             &   30.7               \\
%ResNet-50+MoCo v1~\cite{moco}  &  38.5    &     58.3     &    41.6        &   33.6       &    54.8        &   35.6          \\
Res50+MoCo v2~\cite{mocov2}  &  38.5    &     58.3     &    41.6        &   33.6       &    54.8        &   35.6          \\
Res50+InfoMin~\cite{infomin}  &  39.0    &     58.5     &    42.0        &   34.1       &    55.2        &   \textbf{36.3}          \\
\hline
Res50+CosLr     & 38.2      & 58.2        & 41.2            & 33.3      & 54.7           & 35.2             \\
Res50+CTC(Ours)  & \textbf{39.5}     & \textbf{58.7}  & \textbf{42.0}   & \textbf{34.2}    & \textbf{55.4}   & 36.2   \\ 
%\hline\hline                        
%ResNet-50+AA+CosLr     & 38.4         & \textbf{58.9}    & 41.5          & 32.9      & 54.5           & 35.1             \\
%ResNet-50+AA+CTC(Ours)  & \textbf{39.4}   & 58.7   & \textbf{42.1}       & \textbf{34.1}      & \textbf{55.1}           & \textbf{35.8}            \\ 
\Xhline{2\arrayrulewidth}
%ResNet-50+AA+CosLr+270epoch        & 77.82/xx.xx                    & 71.1                   \\
\end{tabular}
\end{small}
%%\vspace{-10pt}
\caption{COCO object detection and instance segmentation based on Mask-RCNN-FPN with $1\times$ schedule.}
%\vspace{-10pt}
\label{tab:transfer_coco}
\end{table*}

%%\vspace{5pt}
\noindent \textbf{Object detection and instance segmentation. }
%%%\vspace{-0.2cm}
%%%\vspace{-0.3cm}
In this part, we transfer the learned representations of CTC to the object detection and instance segmentation tasks~\cite{rethinkingpretrain1,moco,densecon}. Models pre-trained on ImageNet are further fine-tuned with the Mask-RCNN-FPN~\cite{mask,fpn} on the MS-COCO dataset~\cite{coco} in the commonly applied training protocol.

% \TODO{supp contains the experiment of various arch for object detection and segmentation}
%%%\vspace{-0.3cm} 
Transferring results are summarized in Table~\ref{tab:transfer_coco}. Compared to the vanilla pre-training with cosine learning rate~(CosLr), our CTC yields consistently better transfer performances on the COCO dataset. 
Compared with self-supervised learning methods, \ie, MoCo v2~\cite{mocov2} and InfoMin~\cite{infomin}, CTC also achieves comparable or better results on COCO. For one thing, it indicates that our method also preserves information concerning the detection task. For another, the superiority of SSL in transferring tasks is challenged, \ie, our method proves that supervised pre-training has the potential to achieve better results than SSL methods.

\begin{table}[]
\centering
\begin{small}
\begin{tabular}{l|c|c}
\Xhline{2\arrayrulewidth}
\multirow{2}{*}{pre-training method} &  CUB200  & Aircraft \\
                        & top-1 acc.~(\%)                     &top-1 acc.~(\%)  \\ \hline
Res50+CosLr$\dagger$        & 62.5                      & 27.8               \\
Res50+CTC(Ours)$\dagger$     & \textbf{63.7}             & \textbf{28.2}       \\ 
\hline
Res50+AA+CosLr$\dagger$       & 64.8                   & 31.2       \\
Res50+AA+CTC(Ours)$\dagger$    & \textbf{66.1}          & \textbf{32.1}     \\ 
\hline  
\hline
Res50+CosLr$\ddagger$        & 80.1                      & 82.5               \\
Res50+CTC(Ours)$\ddagger$     & \textbf{81.7}             & \textbf{84.1}       \\ 
\hline
Res50+AA+CosLr$\ddagger$        & 81.3                      & 83.4               \\
Res50+AA+CTC(Ours)$\ddagger$     & \textbf{83.5}             & \textbf{85.6}       \\ 
\Xhline{2\arrayrulewidth}
\end{tabular}
\end{small}
%%\vspace{-10pt}
\caption{Top-1 classification accuracies~(\%) on transferring representations to FGVC datasets. ``$\dagger$'' and ``$\ddagger$'' denote the backbone network is frozen and unfrozen, respectively.}
%\vspace{-20pt}
\label{tab:fine_grained}
\end{table}

%More experiments can be found in our supplementary materials.

%\begin{table}[t]
%\centering
%\begin{scriptsize}
%\begin{tabular}{l|c|c}
%\Xhline{2\arrayrulewidth}
%\multirow{2}{*}{Method} & \multicolumn{2}{c}{Dataset}                                                                                           \\ \cline{2-3} 
%                        & CUB-200                     & Caltech-101 \\ \hline
%ResNet-50+CosLr        & 17.9                      & 67.1               \\
%ResNet-50+AA+CosLr     & 19.2                      & 68.4               \\
%ResNet-50+AA+SupCon\cite{supcon}     & 19.6                      & 68.7               \\
%ResNet-50+AA+CTC(Ours)     & \textbf{20.4}                      & \textbf{69.3}               \\
%\Xhline{2\arrayrulewidth}
%\end{tabular}
%\end{scriptsize}
%%%\vspace{-10pt}
%\caption{mAP~(\%) on the CUB-200 and Caltech-101 datasets.}
%%%\vspace{-10pt}
%\label{tab:image_search}
%\end{table}

%%\vspace{5pt}
\noindent \textbf{Fine-grained visual categorization. }
%%%\vspace{-0.2cm}
%%%\vspace{-0.3cm}
%Our proposed CTC is tailored for visual categorization tasks, it is spontaneous to evaluate the transferability of representations on other classification datasets. 
In this part, representations are transferred to popular fine-grained visual categorization~(FGVC)~\cite{macnn,racnn} datasets. With the backbone network pre-trained on ImageNet frozen and unfrozen, two kinds of transferring experiments are conducted on the CUB-200-2011~(CUB200)~\cite{cub} and the FGVC-Aircraft~(Aircraft)~\cite{aircraft} datasets. To compare with SSL methods, we extend the experiment on the large-scale iNaturalist-18~(iNat-18)~\cite{inat18} dataset. Training details are attached in appendix.
%%%\vspace{-0.3cm}
%\textcolor{red}{Two datasets~(CUB-200-2011 and Aircraft)} are evaluated.
% maybe add an experiment of training CTC directly on FGVC dataset, and put it in supp.

%%%\vspace{-0.3cm}
Transferring results of CUB200 and Aircraft are reported in Table~\ref{tab:fine_grained} and~\ref{tab:inat}. By freezing the backbone parameters and training a linear classifier on top of the learned representations, we observe that the representations learned by CTC achieve better performance over the vanilla baseline. Results of training the model with unfrozen backbone are also provided in Table~\ref{tab:fine_grained}. CTC again outperforms the vanilla training, further validating its effectiveness and practicality. Notably, on iNat-18, our method also outperforms the MoCo v1~\cite{moco}, even the model is pre-trained on the billion-level data Instagram-1B~\cite{wsl}. It further demonsrates that learning transferable representations in the supervised learning is a promising research direction.
%%%\vspace{-0.3cm}

\begin{table}[]
\centering
\begin{small}
\begin{tabular}{l|c}
\Xhline{2\arrayrulewidth}
\multirow{2}{*}{pre-training method}                        & iNat-18        \\ 
 & top-1 acc.~(\%) \\ 
\hline
Res50+CosLr               & 66.1                    \\
Res50+MoCo v1~(IN-1M)~\cite{moco}    & 65.6                   \\
Res50+MoCo v1~(IG-1B)~\cite{moco}    & 65.8                   \\
Res50+CTC~(ours)           & \textbf{66.4}                   \\
\hline
Res50+AA+CosLr            & 66.3                    \\
Res50+AA+CTC~(ours)        & \textbf{66.7}                   \\
\Xhline{2\arrayrulewidth}
\end{tabular}
\end{small}
%%\vspace{-10pt}
\caption{Top-1 accuracies~(\%) on iNaturalist and the backbone is frozen. ``IN-1M'' and ``IG-1B'' denote pre-training with ImageNet-1M~\cite{imagenet} and web Instagram-1B~\cite{wsl} datasets, respectively. All methods (except MoCo v1 IG-1B) are pre-trained on IN-1M. }
%\vspace{-20pt}
\label{tab:inat}
\end{table}

%\TODO{image searching put in supp}
%%%\vspace{5pt}
%\noindent \textbf{Image searching. }
%%%%\vspace{-0.2cm}
%%%%\vspace{-0.3cm}
%Representations with good transferability and discriminability ought to yield satisfactory results on image searching tasks. Hence, we evaluate the image searching problem on the CUB-200-2011~\cite{cub} and the Caltech-101~\cite{caltech} datasets. Directly, representations after the global average pooling are extracted and normalized. Cosine similarities are calculated to measure distances between samples. The mAP results are reported.
%%%%\vspace{-0.5cm}
%
%%%%\vspace{-0.3cm}
%Image searching results are reported in Table~\ref{tab:image_search}. Our CTC consistently yields satisfactory searching accuracies, demonstrating that representations learned by our CTC can generalize on image searching tasks well. We believe the better results are credited to the higher discriminability, which comes from the transferability preservation.

%\section{Limitations}

% 1. two stage，不够简洁；
% 2. 没有找到直接优化I(X;T)的方法；

\section{Discussion and Conclusion}
%The transferability property of representations is often omitted in supervised learning tasks. 
This study focuses on learning representations with good discriminability and transferability at the same time. The trade-off between these properties is firstly observed by us via a temporal analysis. To explain this incompatibility, we explore the correlation between information-bottleneck trade-off and our observed trade-off, and reveal the over-compression phenomenon.
Moreover, we investigate how and why the InfoNCE loss can alleviate the over-compression, and further present the contrastive temporal coding method. Our method successfully make discriminability and transferability compatible. Remarkable transfer learning performances are also achieved. We hope that this work can arouse attentions to the transferability of representations in the conventional supervised learning tasks. In the future, we will explore the existence of over-compression on other popular tasks, \eg, self-supervised learning, object detection and large-scale pre-training. \\

\noindent \textbf{Acknowledgements.}
This work was supported in part by the Zhejiang Provincial Natural Science Foundation of China under Grant No.LQ22F020006. We thank anonymous reviewers from ECCV 2022 for insightful comments.

\newpage

\newcounter{alphasect}
\def\alphainsection{0}

\let\oldsection=\section
\def\section{%
  \ifnum\alphainsection=1%
    \addtocounter{alphasect}{1}
  \fi%
\oldsection}%

\renewcommand\thesection{%
 \ifnum\alphainsection=1% 
   \Alph{alphasect}%
 \else
   \arabic{section}%
 \fi%
}%

\newenvironment{alphasection}{%
  \ifnum\alphainsection=1%
    \errhelp={Let other blocks end at the beginning of the next block.}
    \errmessage{Nested Alpha section not allowed}
  \fi%
  \setcounter{alphasect}{0}
  \def\alphainsection{1}
}{%
  \setcounter{alphasect}{0}
  \def\alphainsection{0}
}%

\begin{alphasection}

\section{Appendix}

The appendix is composed of 7 parts. In Sec.~\ref{subsec:pca}, we discuss the correlation between mutual information maximization and the Principal Components Analysis~(PCA). In Sec.\ref{subsec:more_over_compression}, we provide extensive experiments to reveal the drawback of over-compression. In Sec.~\ref{subsec:more_ctc}, we provide more experiments to validate our CTC alleviates the over-compression. In Sec.~\ref{subsec:overfit}, we provide training and test loss curves to prove that over-compression is not caused by over-fitting. In Sec.~\ref{subsec:mine}, we elaborate the method for estimating mutual information, \ie, MINE~\cite{mine}. In Sec.~\ref{subsec:sec3}, we detail implementation details of all experiments in the manuscript. In Sec.~\ref{subsec:code}, we provide a pseudo code of CTC.

\subsection{Connection between Mutual Information and PCA}
\label{subsec:pca}

PCA is a linear dimensionality reduction method which finds the linear projection(s) of the data that has the maximum variance.
Specifically, given a dataset $\mathcal{D}=\{x_1,\dots,x_n\}$ where $x_n\in\mathcal{R}^D$ with an assumption that the data is zero mean, $\frac{1}{N}\sum_i x_i =0$ and $P(x)$ is Gaussian. Solving for the linear projection $y=w^\top x$ with PCA is solving for the following equation:
$$
w^*=\arg \max_{w} var(y),
$$
where $var(\cdot)$ is the variance function.
Increasing the norm of $w$ increases the variance of $y$, so we limited the norm of $w$ to be a unit, \ie $\|w\|=1$.

Consider we are interested in finding another linear projection $\hat{y}=\hat{w}^\top x$ that has the maximum mutual information $I(x;\hat{y})$. We can have:
$$
I(x;\hat{y}) = H(\hat{y}) - H(\hat{y}|x) = H(\hat{y})
$$
So the goal is to maximize the entropy of $H(\hat{y})$, and because $x$ is a zero mean gaussian, the linear transformation of $x$ is still gaussian. Then, we have:
$$
H(\hat{y}) = -\int p(\hat{y}) \log p(\hat{y}) d\hat{y} = \frac{1}{2} \ln |\Sigma| + \frac{D}{2}(1 + \ln 2\pi),
$$
where $\Sigma$ is the covariance matrix.
Therefore, maximize the entropy of $H(\hat{y})$ is maximize the variance of $\hat{y}=\hat{w}^\top x$ which is solving for:
$$
\hat{w}^* = \arg \max_{\hat{w}} var(\hat{y}) \text{  subject to }\|\hat{w}\| = 1
$$
So solving the PCA and solving for a linear projection that have the maximum mutual information is the same.

\subsection{Observing Over-Compression on More Datasets and Networks}
\label{subsec:more_over_compression}
In this part, we provide more transferring results on various datasets and networks. The source dataset is CIFAR-100~\cite{cifar}, and the target datasets are composed of 6 widely used datasets, \ie, CIFAR-10~\cite{cifar}, STL-10~\cite{stl10}, CINIC-10~\cite{cinic10}, TinyImageNet~\footnote{https://www.kaggle.com/c/tiny-imagenet}, SVHN~\cite{svhn}, and Fashion-MNIST~\cite{fashion_mnist}. Backbone choices are composed of 4 effective models, \ie, ResNet18~\cite{resnet}, WRN-28-4~\cite{wrn}, ShuffleNetV2~\cite{shufflev2}, and ResNext32-16x4d~\cite{resnext}. It could be observed in Figure~\ref{fig:appendix_baseline_transfer} that, with different networks and datasets, the transferability consistently goes to a peak and begins to decrease.

\begin{figure}[h]
\begin{center}
\includegraphics[width=0.95\linewidth]{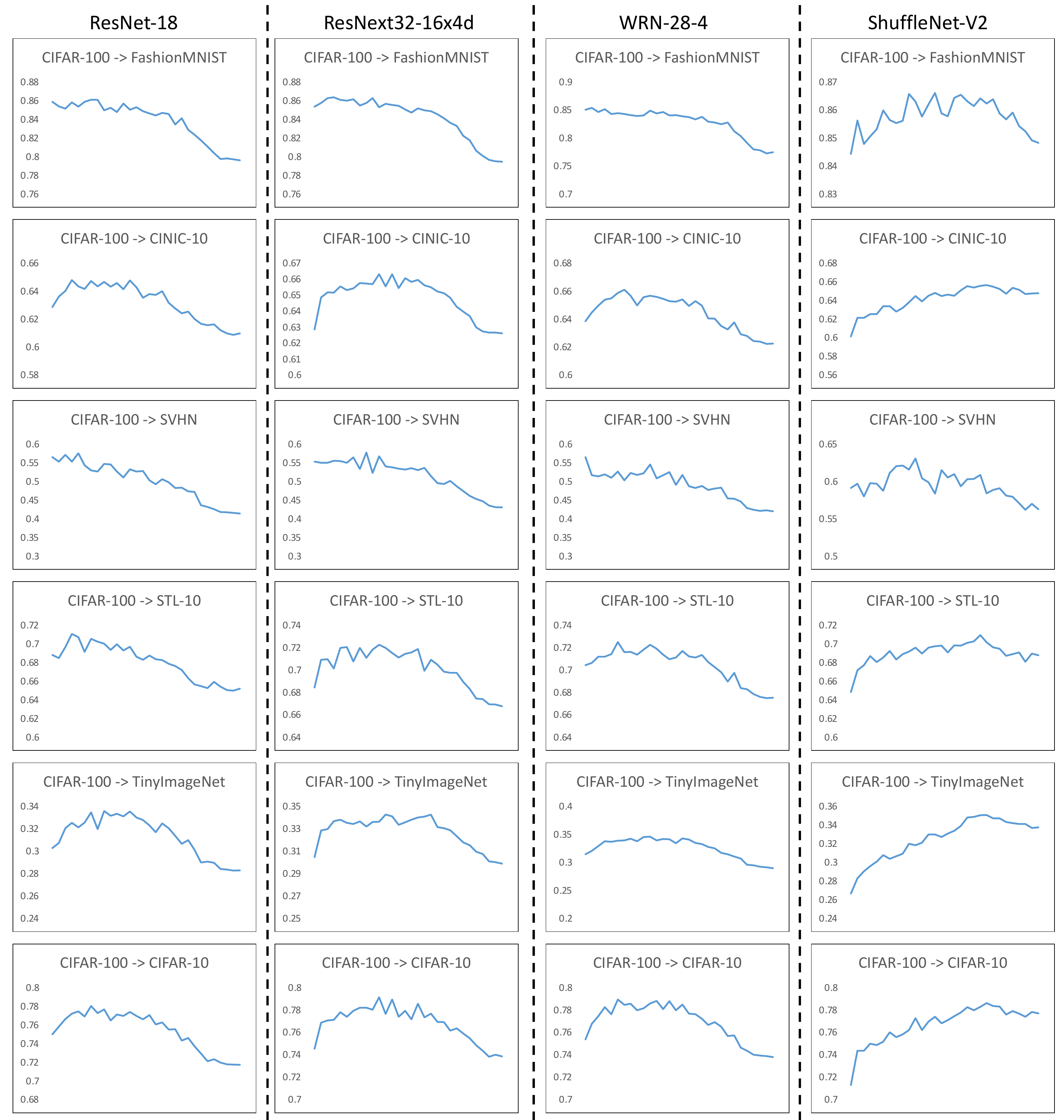}
\end{center}
\vspace{-10pt}
\caption{Temporal transferring results of \textcolor{lineblue}{vanilla} training on the CIFAR-10, STL-10, CINIC-10, TinyImageNet, SVHN, and Fashion-MNIST datasets. X and Y axes represent the training process and top-1 accuracy, respectively. }
\label{fig:appendix_baseline_transfer}
\end{figure}

Similar to experiments in Section~3 of the manuscript, we further provide the information dynamics of the aforementioned networks. Specifically, we evaluate the $I(X;T)$ and $I(T;Y)$ on source dataset~(CIFAR-100) and target datasets~(STL-10 and CINIC-10), with ResNet-18, ResNext32-16x4d, WRN-28-4, and ShuffleNetV2. Results are reported in Figure~\ref{fig:appendix_baseline_info}, we still observe over-compression since $I(X;T)$ decrease on all experiments.

\begin{figure}[htp!]
\begin{center}
\includegraphics[width=0.95\linewidth]{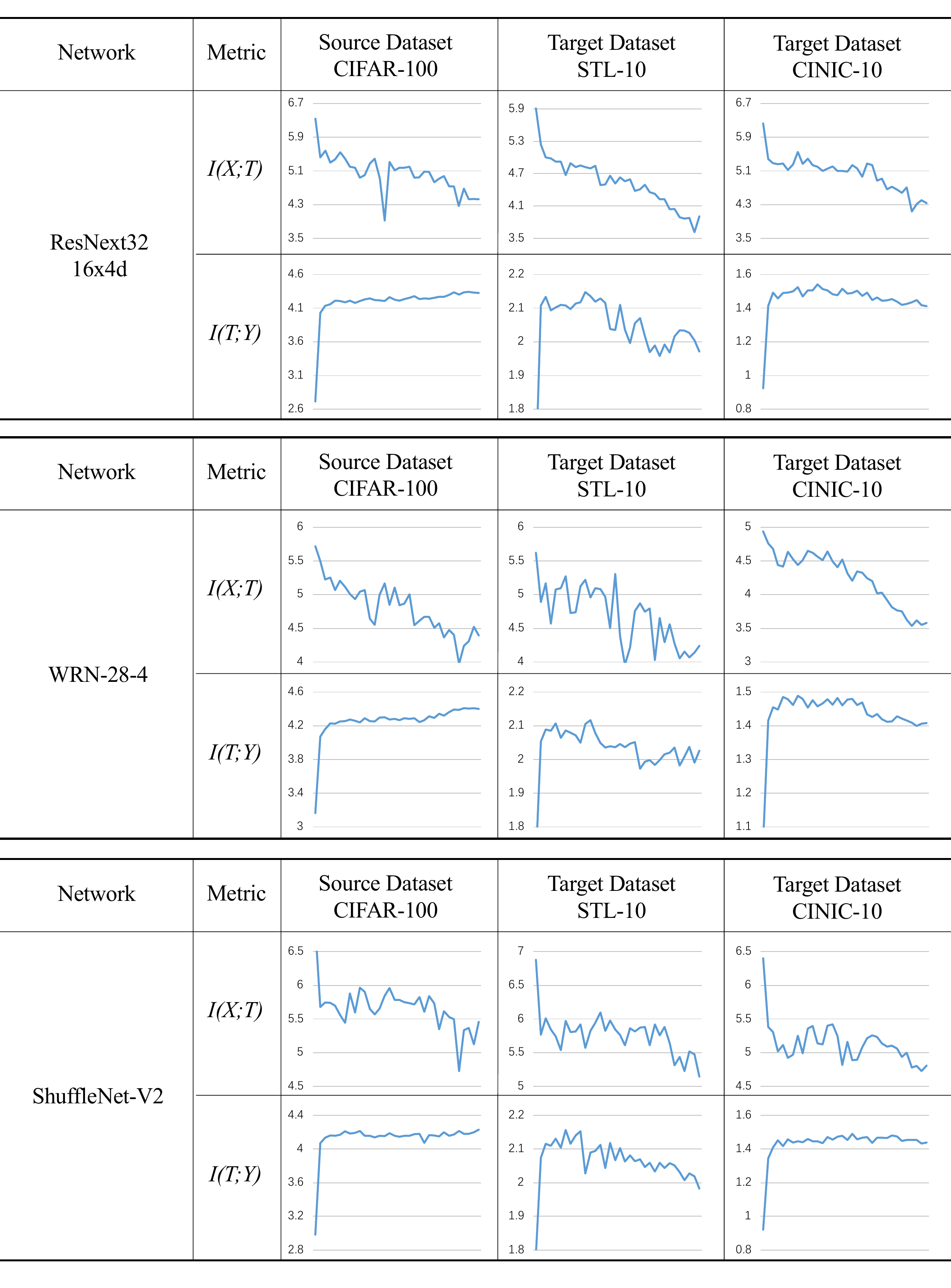}
\end{center}
\vspace{-10pt}
\caption{Mutual information dynamics of \textcolor{lineblue}{vanilla} training on CIFAR-100 and transferring to STL-10 and CINIC-10. }
\label{fig:appendix_baseline_info}
\end{figure}

\newpage

\subsection{Alleviating Over-Compression on More Datasets and Networks}
\label{subsec:more_ctc}
In this part, we demonstrate out proposed CTC alleviates over-compression on all evaluated datasets and networks mentioned in Sec.~\ref{subsec:more_over_compression}. Especially, in the later training, the transferability does not decrease. It can be reflected that over-compression could be a general problem in the transfer learning, and deserves further explorations. Our proposed method provides a practical strategy for achieving this goal.

\begin{figure}[h]
\begin{center}
\includegraphics[width=0.95\linewidth]{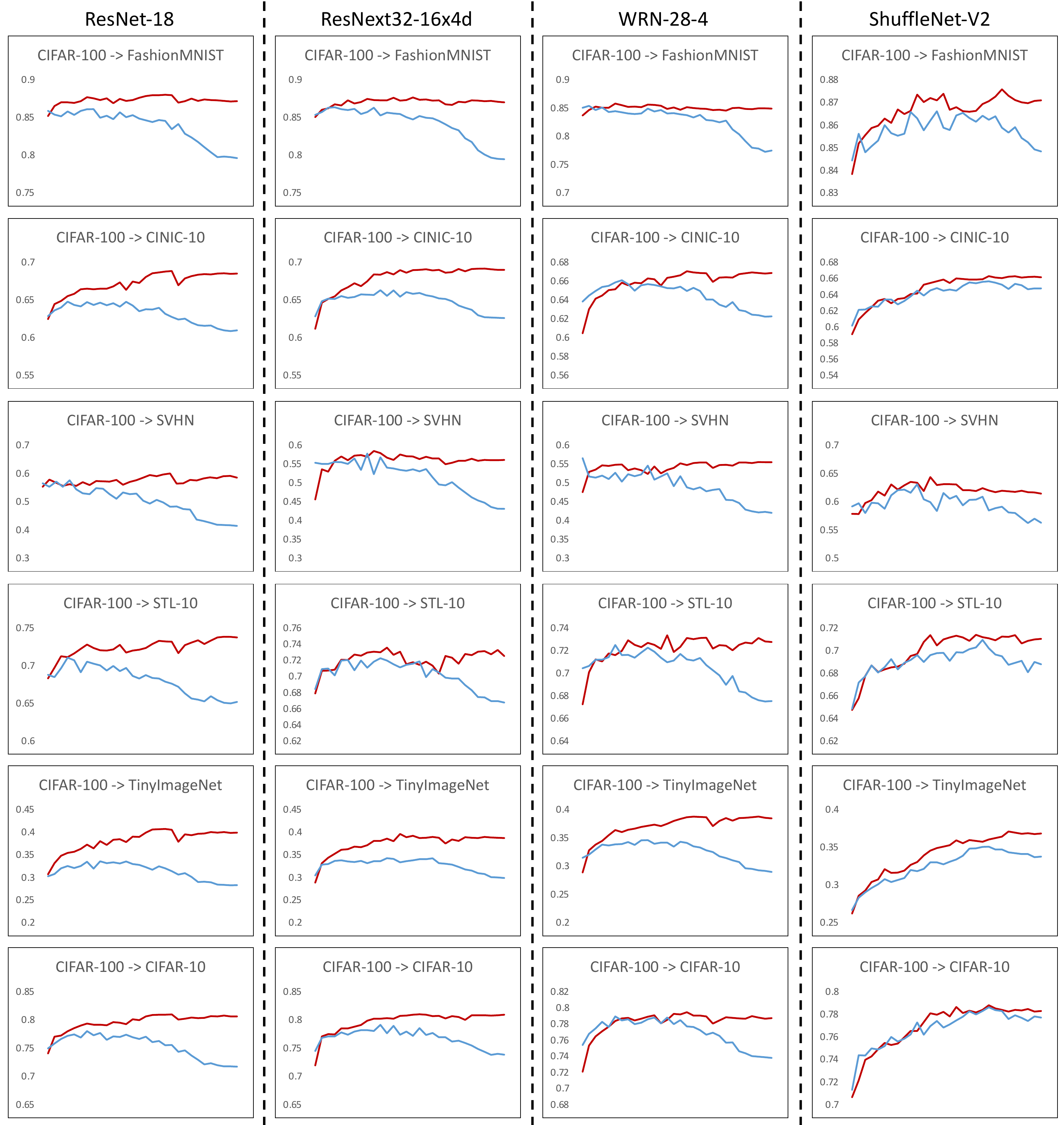}
\end{center}
\vspace{-10pt}
\caption{Temporal transferring results of \textcolor{lineblue}{vanilla} and \textcolor{linered}{CTC} training on the CIFAR-10, STL-10, CINIC-10, TinyImageNet, SVHN, and Fashion-MNIST datasets. X and Y axes represent the training process and top-1 accuracy, respectively. Here we use $\alpha=0.5$ for proving that CTC is able to achieve much better tranferability than the vanilla training. }
\label{fig:appendix_transfer}
\end{figure}

\newpage

\subsection{Does over-fitting happen?}
\label{subsec:overfit}
As defined in Deep Learning book~\footnote{https://www.deeplearningbook.org/contents/ml.html, pp.~108--114.}, over-fitting happens when the generalization error on the \textit{source test} set decreases, but we did not observe such decreases on source datasets. As shown in Figure~\ref{fig:appendix_overfit}, we provide the training and test errors of training ResNet-18 on CIFAR-100, and no decrease of generalization error is observed. Over-compression is related to the deteriorating transferring results while the performance on the source test set keeps improving.

\begin{figure}[h]
\begin{center}
\includegraphics[width=0.95\linewidth]{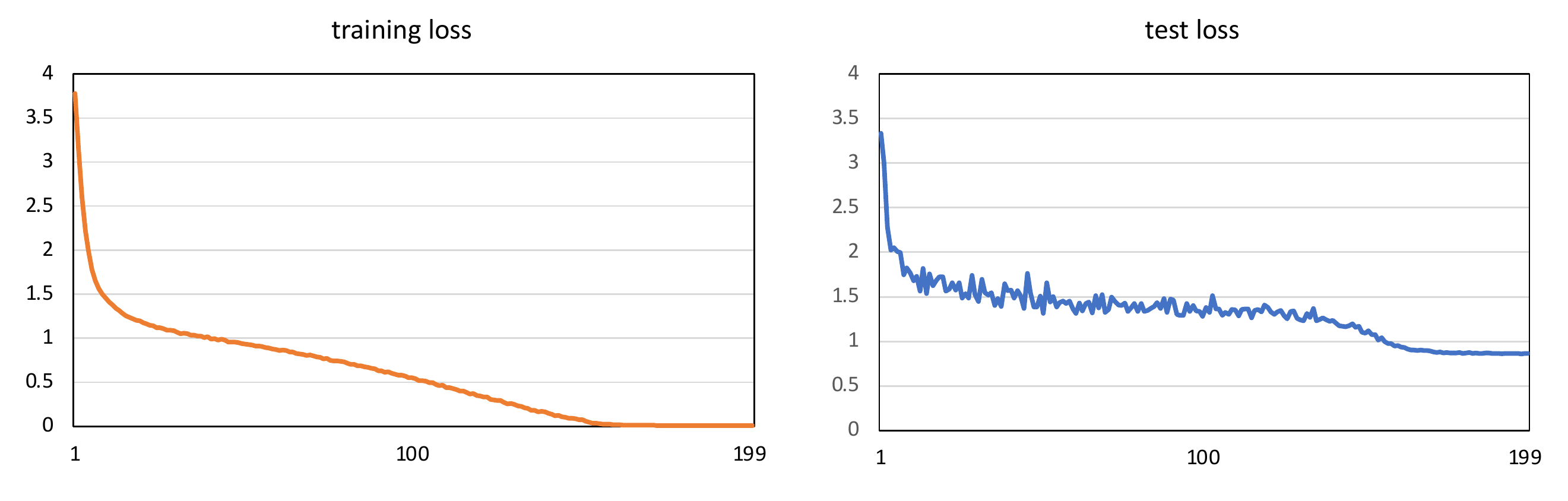}
\end{center}
\vspace{-10pt}
\caption{Training and test loss curves of ResNet-18 on CIFAR-100. }
\label{fig:appendix_overfit}
\end{figure}

\subsection{Mutual Information Neural Estimation}
\label{subsec:mine}

Mutual Information Neural Estimation~(MINE)~\cite{mine} estimates mutual information $I(X;Y)$ by training a classifier to distinguish between samples from the joint, $\mathbb{J}$, and the product of marginals, $\mathbb{M}$, of random variables $X$ and $Y$. 

We implement our estimator based on the open-source code from GitHub~\footnote{https://github.com/sungyubkim/MINE-Mutual-Information-Neural-Estimation-}.
In our implementation, we adopt a Multi-Layer Perceptron~(MLP) composed of four fully-connected layers with hidden dimension 1024, and ReLU is used as the activation function. 
For calculating $I(X;T)$, the input dimension is set to $3072~(32\times 32\times 3) + 512$, which is the summation of the resolution of tiny images and the dimension of representations. The batch size and learning rate is set to 1K and 1e-4, respectively. For each model, we train the MLP for 10K steps. For calculating $I(T;Y)$, the input dimension is $512 + C$, which represents the summation of the dimention of representations and the number of classes in the dataset. The batch size and learning rate is set to 5K and 1e-5, respectively. For each model, we train the MLP for 10K steps. The Adam~\cite{adam} optimizer is used.

\subsection{Implementation Details}
\label{subsec:sec3}
\textbf{In Sec.~3 of the manuscript}, we develop the baseline used for temporal analyses . 

\vspace{5pt}
\noindent \textbf{Sec.~3.1: Training on CIFAR-100. } We optimize a ResNet18~\cite{resnet} on the source dataset CIFAR-100~\cite{cifar}, with the SGD optimizer and a batch size of 64. The initial learning rate is 5e-2, and the learning rate follows a cosine decay scheduler. The weight decay is set to 5e-4 and the total training epoch is set to 200. For avoiding effects of over-fitting on information dynamics, we set a minimum learning rate 1e-2 and early stop the training at the 190-th epoch.

\vspace{5pt}
\noindent \textbf{Sec.~3.1: Transferring to STL-10 and CINIC-10. } For STL-10~\cite{stl10} and CINIC-10~\cite{cinic10} datasets, we re-train a classifier on top of the backbone learned on the source dataset at each epoch. Specifically, for every evaluated model, we train the classifier for 15K steps. We set the batch size to 512 and the initial learning rate to 4e-1. The learning rate is decayed by 0.1 at the 5K-th and 10K-th steps, respectively.

% \subsection{Implementation Details in Sec.~5.}

% \noindent \textbf{Experiments in Sec.~5.1. } 
\vspace{10pt}
\textbf{In Sec.~5.1 of the manuscript}, we prove our motivation on CIFAR-100, and then conduct image classification tasks on both CIFAR-100 and ImageNet~\cite{imagenet}.

\vspace{5pt}
\noindent \textbf{Sec.~5.1: Benchmarking on CIFAR-100.} We mainly follow the baseline settings in Sec.~\ref{subsec:sec3}. Specifically, we set the initial learning rate to 5e-2 and the batch size to 64. The SGD optimizer and cosine learning rate scheduler are used. Note that we optimizer the model for 300 epochs as the baseline for fair comparisons with our CTC. As settings about CTC, we optimize the first stage~(information aggregation stage) for 200 epochs and the second stage~(information revitalization stage) for 100 epochs. The initial learning rate of the second stage is set to 5e-3 and also follows a cosine scheduler. The $\alpha$ and $\beta$ are set to $0.01$ and $1.0$, respectively.

\vspace{5pt}
\noindent \textbf{Sec.~5.1: Benchmarking on ImageNet. } For experiments on the ImageNet dataset, we set the initial learning rate to 0.01 with a batch size of 256. We use the SGD optimizer and the cosine learning rate scheduler. The model is trained for a total of 200 epochs, where 120 epochs are for the first stage and the last 80 epochs are for the second stage.
The parameters $\alpha$ and $\beta$ are set to $0.2$ and $1.0$ respectively.
% \vspace{5pt}
% \noindent \textbf{Experiments in Sec.~5.2. } 

\vspace{10pt}
\textbf{In Sec.~5.2 of the manuscript}, we introduce the AutoAugment~\cite{autoaug} and plug in our CTC for better transferability.

\vspace{5pt}
\noindent \textbf{Sec.~5.2: Benchmarking on CIFAR-100.} We set the initial learning rate to 5e-2 and the batch size to 64. The SGD optimizer and cosine learning rate scheduler are used. Note that we optimizer the model for 300 epochs as the baseline for fair comparisons with our CTC. As settings about CTC, we optimize the first stage~(information aggregation stage) for 200 epochs and the second stage~(information revitalization stage) for 100 epochs. The temperature parameters for the first and second stages are 0.5 and 0.4. The $\alpha$ and $\beta$ are set to $0.1$ and $1.0$, respectively.

\vspace{5pt}
\noindent \textbf{Sec.~5.2: Benchmarking on ImageNet. } The experiments on ImageNet with AutoAugment are the same with Sec.~5.1.

\vspace{10pt}
\textbf{In Sec.~5.3 of the manuscript}, we transfer representations to various tasks, \ie, object detection on COCO~\cite{coco} and fine-grained visual categorization~(FGVC) on CUB200~\cite{cub}, Aircraft~\cite{aircraft}, and iNaturalist-18~\cite{inat18}. 

\vspace{5pt}
\noindent \textbf{Sec.~5.3: Object detection on COCO. }
For the experiments transferring the learned representation to object detection on COCO, we use the \texttt{train2017} split for training the model and the \texttt{val2017} split to test the finetuned model.
We adopted the Mask-RCNN~\cite{mask} with FPN~\cite{fpn} as the architecture for detection, and the model is trained with the $1\times$ schedule with a maximum of 180K iterations of training, the learning rate is set to 0.02 and the batchsize is 16, step decay schedule is used, the learning rate will be multiplied by 0.1 at 120K and 160K iterations.

% \vspace{5pt}
% \noindent \textbf{Fine-grained visual categorization in Sec.~5.3.}

\vspace{5pt}
\noindent \textbf{Sec.~5.3: FGVC on CUB200, Aircraft, and iNaturalist-18. }
For transfer learning experiments on fine-grained classification datasets, we finetune the pretrained model with 100 epochs, and the learning rate is set to 5e-3 with cosine decay, and the batchsize is 64 for CUB-200 and Aircraft, and is 256 for iNaturalist-18.

%\vspace{5pt}
%\noindent \textbf{Sec.~5.3: FGVC on iNaturalist-18}

\begin{algorithm*}[t]
    \caption{Pseudo code of CTC in a PyTorch-like style.}
    \label{algo:ctc}
    \scriptsize
    \begin{alltt}
    \color{OliveGreen}
    # net: the network 
    # memory: the memory bank for holding representations
    # e1, e2: numbers of epochs for two stages
    # extract: a function to extract representations
    # sample: a function to sample representations from the memory bank
    # alpha, beta: hyper-parameters
    \end{alltt}
    \vspace{-15pt}
    
    \begin{alltt}
   	\color{magenta}for\color{Black} _ \color{magenta}in\color{Black} e1: \color{OliveGreen}# Information aggregation stage \color{Black}
      \color{magenta}for\color{Black} x \color{magenta}in\color{Black} loader:
        logits = net.forward(x)
        t_1 = net.extract(x)
        v_1 = memory.sample(x) \color{OliveGreen}# Sampling contrastive samples from the memory \color{Black}
        loss_ias = ContrastiveLoss(t_1, v_1)\color{OliveGreen}# Calculating the IAS loss \color{Black}
        loss_ce = CrossEntropyLoss(logits, labels) 
        loss = alpha * loss_ias + loss_ce
        loss.backward()
        update(net.param)
        update(memory, t_1) \color{OliveGreen}# Updating memory bank \color{Black}
        
    information_bank = net

    \color{magenta}for\color{Black} _ \color{magenta}in\color{Black} e2: \color{OliveGreen}# Information revitalization stage \color{Black}
      \color{magenta}for\color{Black} x \color{magenta}in\color{Black} loader:
        logits = net.forward(x)
        t_2 = net.extract(x)
        t_1_hat = information_bank.extract(x).detach()
        loss_irs = ContrastiveLoss(t_2, t_1_hat) \color{OliveGreen}# Calculating the IRS loss \color{Black}
        loss_ce = CrossEntropyLoss(logits, labels)
        loss = beta * loss_irs + loss_ce
        loss.backward()
        update(net.param)
	\end{alltt}
\end{algorithm*}

\subsection{Pseudo Code of CTC}
\label{subsec:code}

A PyTorch-Style pseudo code of the CTC method is given in Alg~\ref{algo:ctc}.

\end{alphasection}

\bibliographystyle{splncs04}
\bibliography{egbib}
\end{document}